\documentclass{article}

\usepackage{arxiv}

\usepackage[utf8]{inputenc} % allow utf-8 input
\usepackage[T1]{fontenc}    % use 8-bit T1 fonts
\usepackage{url}            % simple URL typesetting
\usepackage{booktabs}       % professional-quality tables
\usepackage{amsfonts}       % blackboard math symbols
\usepackage{nicefrac}       % compact symbols for 1/2, etc.
\usepackage{microtype}      % microtypography
\usepackage{graphicx}
\graphicspath{ {./images/} }

\newcommand{\etal}{\textit{et al.}}
\newcommand{\ie}{\textit{i.e.}}
\newcommand{\eg}{\textit{e.g.}}

\usepackage{amsmath}
\usepackage{amssymb}
\usepackage{booktabs}
\usepackage{multirow}
\usepackage{caption}
\usepackage{subcaption}
\usepackage{array}
\usepackage{tabularx}

\title{Edge2Vec: A High Quality Embedding\\for the Jigsaw Puzzle Problem}

\author{
    Daniel Rika \\
    Dept.~of Computer Science\\
    Bar-Ilan University\\
    Israel, Ramat-Gan 52900 \\
    \texttt{danielrika@gmail.com}
    \And
    Dror Sholomon \\
    Dept.~of Computer Science\\
    Bar-Ilan University\\
    Israel, Ramat-Gan 52900 \\
    \texttt{dror.sholomon@gmail.com}
    \And
    Eli (Omid) David \\
    Dept.~of Computer Science\\
    Bar-Ilan University\\
    Israel, Ramat-Gan 52900 \\
    \texttt{mail@elidavid.com}
    \And
    Nathan S.~Netanyahu \\
    Dept.~of Computer Science\\
    Bar-Ilan University\\
    Israel, Ramat-Gan 52900 \\
    \texttt{nathan@cs.biu.ac.il}
}

\begin{document}
\maketitle
%===========================================================
\begin{abstract}
Pairwise compatibility measure (CM) is a key component in solving the jigsaw puzzle problem (JPP) and many of its recently proposed  variants. With the rapid rise of deep neural networks (DNNs), a trade-off between performance (\ie, accuracy) and computational efficiency has become a very significant issue. Whereas an end-to-end DNN-based CM model exhibits high performance, it becomes virtually infeasible on very large puzzles, due to its highly intensive computation. On the other hand, exploiting the concept of embeddings to alleviate significantly the computational efficiency, has resulted in degraded performance, according to recent studies. This paper derives an advanced CM model (based on modified embeddings and a new loss function, called \textit{hard batch triplet loss}) for closing the above gap between speed and accuracy; namely a CM model that achieves SOTA results in terms of performance and efficiency combined. We evaluated our newly derived CM on three commonly used datasets, and obtained a reconstruction improvement of 5.8\% and 19.5\% for so-called Type-1 and Type-2 problem variants, respectively, compared to best known results due to previous CMs.
\end{abstract}

%===========================================================
\section{Introduction}
% --------- JPP Background --------- %
The jigsaw puzzle problem (JPP) is an ongoing  challenging task, which has been studied for decades.
Although the problem plays a role in various  \textit{real-world} applications, e.g., broken pottery, ancient frescoes, shredded documents, etc., most of the effort has focused on its \textit{synthetic} version of putting together a ``perfect'' digital image cropped into $N$ square non-overlapping pieces, each of size $S \times S$ pixels.
The synthetic puzzle variant has served as a common testbed for many researchers, and has evolved dramatically over the years.
In principle, reconstructing a puzzle requires two main phases: (1) Computation of all pairwise compatibility measures (CMs) between pieces, and (2) employment of a reconstruction algorithm based on the first phase.
The above framework is not only reasonable, but has proven immensely effective in various problem instances, involving, e.g., puzzles with tens of thousands of pieces, puzzles with missing pieces and even multiple puzzles with mixed pieces.
This paper focuses mainly on the first phase, i.e., finding the CM, which directly affects the quality of the resulting reconstruction.

% --------- Classical CM Background --------- %
One of the first CMs is the \textit{sum square differences} (SSD) introduced by Cho et al.~\cite{conf/cvpr/ChoAF10}, which computes the sum of squared (pixel value) differences along the common edge between adjacent pieces. Pomeranz et al.~\cite{conf/cvpr/PomeranzSB11} later proposed the \textit{prediction-based compatibility} (PBC) measure, which tries to quantify how well a left piece edge can ``predict'' an adjacent right piece edge and vice versa. Specifically, they considered a 2-pixel width edge for each adjacent piece of a pair in question, and assessed the prediction by the $(L_{3/10})^{1/16}$ normalization.
Next, Gallagher~\cite{gallagher2012jigsaw} proposed the \textit{Mahalanobis gradient compatibility} (MGC), which compares the difference between color gradients, rather than the difference between RGB values.
Paikin et al.~\cite{paikin2015solving} suggested an asymmetric CM based on the $L_1$ metric (referred to as $L_1$ CM).
The above traditional CMs rely solely on content information of piece edge, and they tend to be computed easily.

% --------- DNN-based CM Background --------- %
Under slightly more realistic conditions, however, of degraded puzzles with eroded piece boundaries, all of the above CMs become virtually useless, since edge pixels along the boundary of adjacent pair pieces lack essential (chromatic) information.
A whole new set of CMs, based on \textit{deep neural network} (DNN) architectures, has emerged, to tackle, e.g., eroded piece boundaries. These DNN-based CMs work as an end-to-end (E2E) function $g:\mathbb{R}^{S \times 2S \times C} \to \mathbb{R}$, as it inputs a pair of pieces of size $S \times S$ and $C$ color channels, and outputs a compatibility score. Since these DNN-based E2E functions arrive at a compatibility score based on the \textit{entire} pair of pieces, they are pretty robust to pieces with eroded boundaries and possibly other forms of degradation.
Pai\~{x}ao et al.~\cite{paixao2018deep} used SqueezeNet~\cite{SqueezeNet} and MobileNetV2~\cite{mobilenetv2_2018_cvpr} as  DNN-based E2E CMs to achieve state of the art reconstruction for \textit{strip-cut} shredded documents.
Rika et al.~\cite{rika2019gecco} designed a DNN-based E2E CM for reconstructing Portuguese tile panels with hundreds of tiles.
Bridger et al.~\cite{bridger2020cvpr} used a GAN-based CM to solve puzzles with eroded boundaries, by first filling the gap between a pair of pieces by the \textit{generator}, and then providing a compatibility score by the \textit{discriminator}.

% --------- Motivation to Embeddings --------- %
In terms of complexity, there are $4N(N-1)$ and $16N(N-1)$ pairwise combinations for the so-called \textit{Type-1} and \textit{Type-2} variants, respectively, of an $N$-piece puzzle. (Type-1 refers to unknown piece location and known orientation, whereas Type-2 refers to unknown piece location and orientation.)
Obviously, using a DNN-based E2E CM, which requires millions (or even billions) of MACs for a single inference, could become extremely expensive computationally, which only worsens as the number of pieces gets larger.

% --------- TEN: Twin Embedding Networks --------- %
To alleviate the above computational burden, a more efficient scheme was proposed by Rika \etal~\cite{ten2022}, called TEN, which works as follows. In contrast to a typical DNN-based E2E model, which computes the CM of a given pair explicitly (\ie, by intensive processing of all pairwise pairs), TEN first represents the given pieces $k_l, k_r \in \mathbb{R}^{S \times S \times C}$ in latent space $z_l, z_r \in \mathbb{R}^d$, using two embedding DNN-based models, $f_{left}$ and $f_{right}$.
These latent representations extract the most relevant information of a tile with respect to its boundary, \ie $f_{left}$ extracts information from the left piece with respect to its right edge, and $f_{right}$ extracts information from the right piece with respect to its left edge.
After obtaining all of the $4N$ boundary representations by $f_{left}$ and $f_{right}$, TEN employs a simple and much faster distance measure between embedding vectors, $\mathcal{D}: \mathbb{R}^d \times \mathbb{R}^d \to \mathbb{R}$, to compute all of the $16N(N-1)$ pairwise CMs. (Euclidean distance was reported to work best.)
Despite of its significantly improved efficiency (relatively to a DNN-based E2E), TEN was even slightly more accurate than classical CMs. However, its lower accuracy compared to that of a powerful DNN-based E2E, poses a substantial gap in performance to be closed, limiting TEN to a mere concept.

% --------- Edge2Vec --------- %
This paper introduces Edge2Vec, a fast, robust, and accurate CM scheme. As opposed to TEN, Edge2Vec eliminates the notion of twin embedding networks, by using the same model to extract  features from both the left and right pieces due to a simple \textit{horizontal flipping} technique.
Furthermore, we demonstrate the superior performance of Edge2Vec relatively to a DNN-based E2E CM in the domain of \textit{eroded boundaries}, which serves as a testbed for real-world settings.
The main contributions of our work are as follows:
\begin{itemize}
    \item[$\bullet$] Closed entirely the performance gap from TEN to a DNN-based E2E CM, without compromising computational efficiency.
    \item[$\bullet$] Proposed a grouping technique to reduce footprint from 9.6M parameters to only 2.1M (\ie $\times$4.6 less parameters) without affecting the performance.
    \item[$\bullet$] Proposed a new variant of triplet loss, called \textit{hard-batch-triplet loss} for high quality edge representation.
\end{itemize}

% TODO
% --------- Outline of the paper --------- %

\section{Edge2Vec Architecture}

We show how to improve TEN in two aspects. First, we eliminate the redundancy of twin embedding models (instead of using a single one), and secondly we propose a technique to handle efficiently the last projection layer that contains the vast majority of the parameters of the model.

\subsection{From Twins to a Single Embedding Network}
\label{sec:twins_to_single}
\begin{figure}
        \centering
        \begin{subfigure}{0.8\linewidth}
        \centering
        \includegraphics[width=\linewidth]{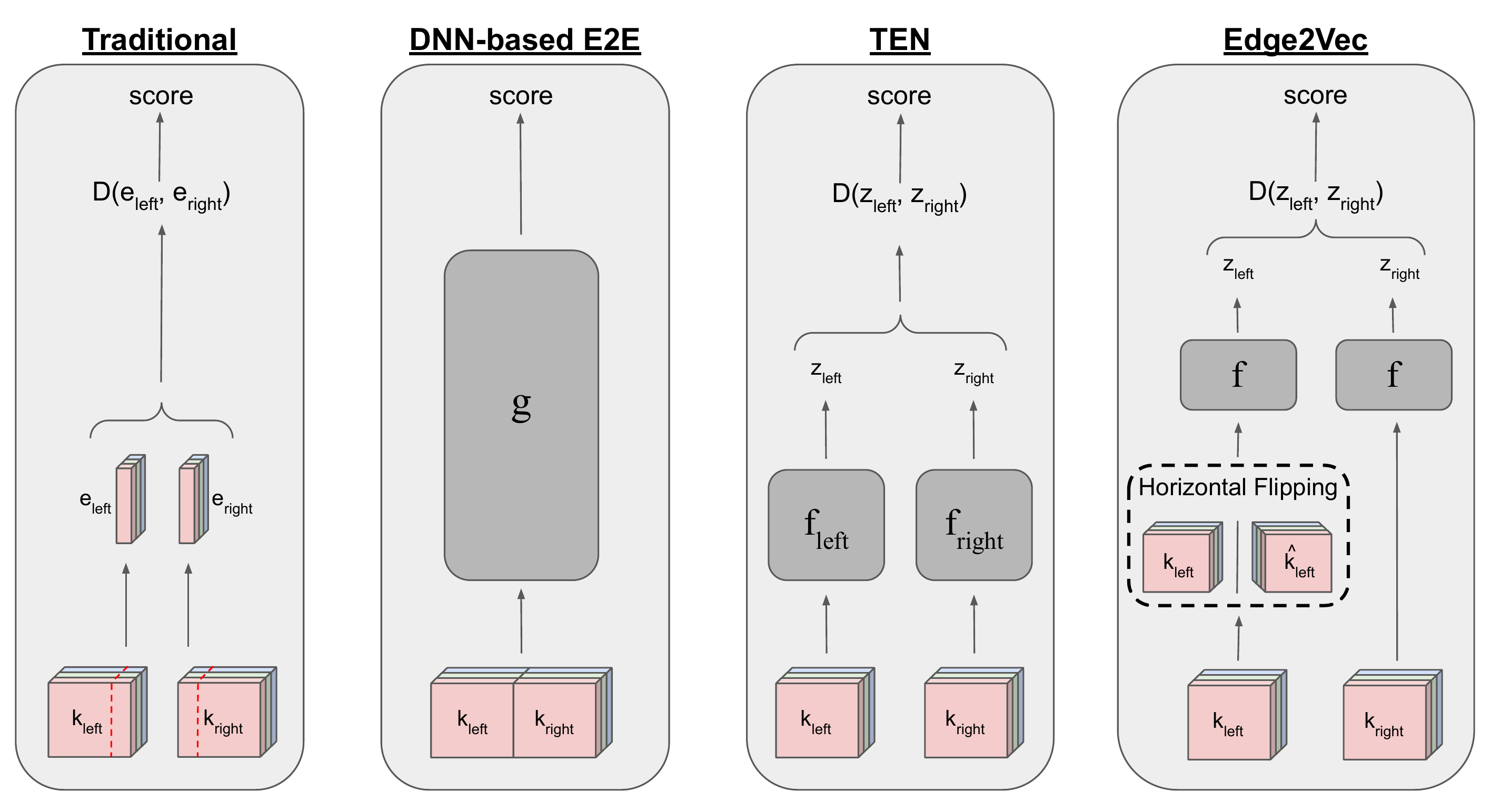}
        \caption{}
        \label{fig:L2_reg:a}
        \end{subfigure}
        % \hfill
        \vspace{0.5cm}
        \\
        \begin{subfigure}{0.7\linewidth}
        \centering
        \includegraphics[width=\linewidth]{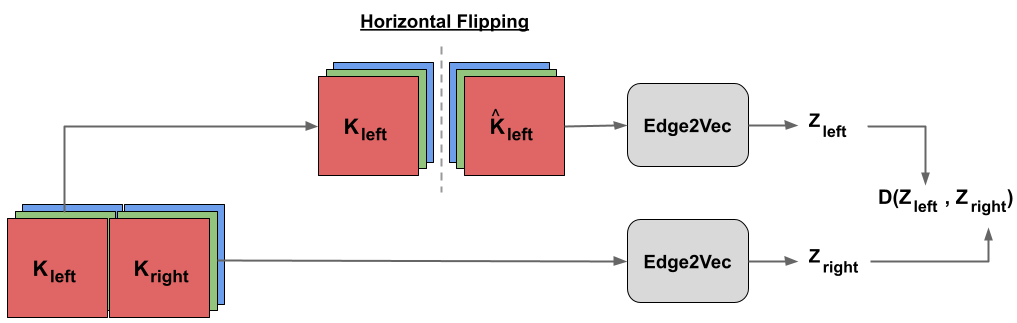}
        \caption{}
        \label{fig:L2_reg:b}
        \end{subfigure}
        \caption{Illustration of pairwise CM schemes: (a) Block diagrams of traditional, E2E, TEN, and Edge2Vec schemes, where $\mathcal{D}$ denotes any distance metric, \eg, Euclidean distance, in our case ; (b) \textit{horizontal flipping} used by Edge2Vec.}
        \label{fig:sen_scheme}
    \end{figure}

    Given a pair of pieces $(k_l, k_r)$, where $k_l, k_r \in \mathbb{R}^{S \times S \times C}$ are the left and right piece in the pair, respectively, TEN first embeds $k_l, k_r$ in latent vectors $z_l, z_r \in \mathbb{R}^d$, and then computes the Euclidean distance between these embeddings, as the CM of the pieces.
    Since $z_l$ represents the information of $k_l$ with respect to its right edge, and $z_r$ represents the information of $k_r$ with respect to its left edge, two different models, $f_{left}, f_{right}$ were proposed.
    We suspect that training and employing such twin models might prove too hard and unstable a task, since each model needs to adjust millions of parameters for generating embedding vectors of the same distribution as the vectors of its companion model with extreme precision.
    Instead, applying \textit{horizontal flipping} (\ie, mirroring) to $k_l$ and $k_r$, thereby getting $\hat{k}_l$ and $\hat{k}_r$, respectively, should imply naturally that $f_{left}(k_l) = f_{right}(\hat{k_l})$ and $f_{right}(k_r) = f_{left}(\hat{k_r})$. For example,  $\hat{k_l}$ contains the same information as $k_l$, only that
    due to the flipping, the right edge of $k_l$ becomes the left edge of $\hat{k}_l$. (A similar observation holds for $k_r$.)
    Thus, the redundancy of TEN's $f_{left}$ and $f_{right}$ models should be clear. Instead, one can represent both left and right edges of a pair of pieces, due to \textit{horizontal flipping}, by a single model, \eg, $z_l=f_{right}(\hat{k}_l)$, $z_r=f_{right}(k_r)$.
    Shifting away from twin embedding networks to a single network, according to the above explanation, is illustrated in Figure~\ref{fig:sen_scheme}b. We will refer to this proposed scheme as Edge2Vec. 
    
\begin{figure}
        \centering
        \includegraphics[width=0.8\linewidth]{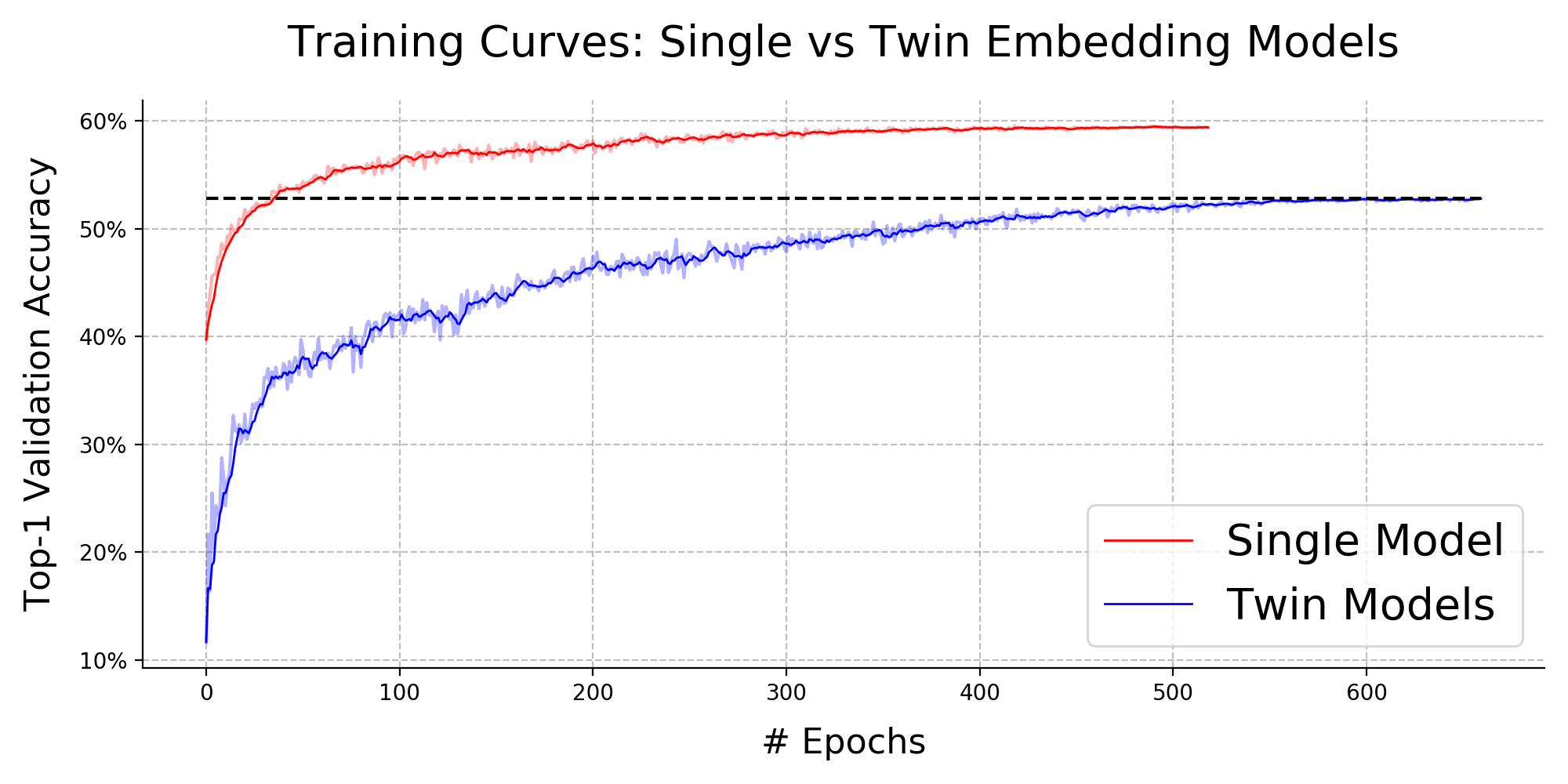}
        \caption{Plots of Top-1 validation accuracy;  single embedding model takes only 38 epochs to reach TEN's highest level, obtained after 600 epochs (see black dashed line); also, single model improves Top-1 validation accuracy by 6.7\%, while using only half the number of TEN's weights (\ie, 2.5M parameters instead of 5.1M).}
        \label{fig:edge2vec_vs_ten_curve}
    \end{figure}

    To justify the use of a single model instead of twins, we trained Edge2Vec on the same configurations and datasets described in \cite{ten2022}. Training plots of Top-1 validation (to be defined) for both models are depicted in Figure~\ref{fig:edge2vec_vs_ten_curve}.
    In addition to its faster convergence -- Edge2Vec needed 15.8 times less epochs to reach TEN's highest Top-1 level -- it also improved TEN's Top-1 accuracy on the test set by 5.7\% and 5.8\% for Type-1 and Type-2, respectively. 

    % \begin{figure}
    %     \centering
    %     \includegraphics[width=\linewidth]{figures/Edge2Vec_scheme.png}
    %     \caption{Illustration of pairwise CM scheme using \textit{horizontal flipping} and Edge2Vec; $\mathcal{D}$ represents any distance metric, \eg \textit{Euclidean distance}, in our case.}
    %     \label{fig:sen_scheme}
    % \end{figure}
    
\subsection{Grouping}
\label{sec:grouping}
    As previously mentioned, Edge2Vec originates from TEN's architecture, which consists of four convolutional layers corresponding to $64, 128, 256$, and $512$ output channels, using a kernel size of $3\times3$ and the nonlinear ReLU activation function~\cite{nair2010rectified}
    after each layer. Max pooling was applied after the second and third convolutional layers to reduce spatial dimensionality, and a fully-connected (FC) layer was finally applied to get the embedding vector. It is well known that the largest number of connections (or weights associated with them) lie between the last convolutional layer and the embedding vector. Specifically, this number of parameters is given by $C_{o}*S_{o}^2*d$, where $d$ is the embedding dimension, $C_{o}=512$ is the number of output channels, and $S_{o}=7$ output spatial size (assuming $28 \times 28$ input images).
    To mitigate such parameter explosion, we found that splitting the FC layer into $G$ groups of smaller FC sub-layers (for certain $G$ values) could downsize the number of parameters to $(C_{o}*S_{o}^2*d)/G$, without suffering  any degradation in performance.
    Before flattening the output of the last convolutional layer $x$, we split it along the channel axis into $G$ sub-channels, $\{x_1, ..., x_{G}\}$. Having defined also $G$ FC sub-layers: $FC_i: \mathbb{R}^{(C_{o}/G) * S_o^2} \to \mathbb{R}^{d/G}$, we then assign each $x_i$ to its own $FC_i$ sub-layer, to obtain $G$ sub-embeddings $\{z_1, ..., z_{G}\}$ whose concatenation provides the final embedding vector $z$. Namely, instead of a single FC layer, we use $G$ FC sub-layers, each of which is associated with only $1/G^2$ the number of parameters. Thus, the above grouping decreases the total number of parameters associated with the full FC layer by a factor of $G$.

\section{Training Edge2Vec}
In order to train Edge2Vec, we started with the \textit{triplet-loss}:
\begin{equation}
    \mathcal{L}_{\text{Triplet}} = \max(0, \mathcal{D}(z_a, z_p) - \mathcal{D}(z_a, z_n) + \gamma)
\end{equation}
where $\mathcal{D}$ denotes Euclidean distance, and $\gamma$ is a hyperparameter set to $1$. We discuss here the choice of a  negative embedding $z_n$, which competes with the positive embedding $z_p$ for the anchor $z_a$.
During training, a triplet $(z_a, z_p, z_n)$ is picked randomly from the same image (which is  also selected randomly from a pool of training images).
It is reasonable to assume that picking $z_n$ from the same image would improve the model's  discrimination capability, since a given image tends to be composed of certain color and texture distributions.
However, this is only one possibility for selecting $z_n$; other strategies might push the model even harder to extract better embedding features, by picking more challenging $z_n$'s from other training set images, to distinguish from.

In addition, during each training iteration, a batch of $B$ triplet samples are picked randomly and stacked together to get more informative gradients for the next optimization step of the model.
Hence, we propose an efficient way which takes advantage of batching for finding harder $z_n$'s.
Our proposed selection of \textit{hard batch triplets} (HBT) is explained, in detail, in the following subsection.

\subsection{Hard Batch Triplet (HBT) Selection}
\label{sec:hard_batch_triplet_loss}
    
    Let us denote each triplet in the batch $(z_a^b, z_p^b, z_n^b)$, where $b \in \{ 1,..,B \}$ is the batch index.
    Instead of looking at each triplet $(z_a^b, z_p^b, z_n^b)$ in the batch separately, we can combine the triplets to obtain a more diverse set of negative candidates, by sharing both the positives and negatives between all of them.
    In other words, every anchor $z_a^b$ will have $2B$ candidates $\tilde{\mathcal{Z}}^b = \{ z_p^1, z_p^2, ..., z_p^B, z_n^1, z_n^2, ..., z_n^B \}$, where $z_p^b$ is the only positive embedding, and all other $2B-1$ embeddings are negative candidates.
    Thus, instead of using $z_n^b$ as the negative for anchor $z_a^b$, we can calculate the distances between $z_a^b$ to all of the $2B-1$ negative candidates in $\tilde{\mathcal{Z}}^b$, and choose the ``hardest'' one in the batch. Put
    formally, for every anchor $z_a^b$ in the batch, its new negative is defined as
    \begin{equation}
         z_{n^*}^b = \operatorname*{argmin}_{z \in \tilde{\mathcal{Z}}^b \backslash \{ z_p^b \} } \mathcal{D}(z_a^b, z)
    \end{equation}
    for which we then apply the regular \textit{triplet loss}:
    \begin{equation}
        \mathcal{L}_{\text{HBT}}(z_a^b, \tilde{\mathcal{Z}}^b) = \max(0, \mathcal{D}(z_a^b, z_p^b) - \mathcal{D}(z_a^b, z_{n^*}^b) + \gamma)
    \end{equation}
    Experimenting with our \textit{HBT loss function} during training, we observed initially a boost in Top-1 performance. However, this improvement quickly tapered off on the validation set, as a result of \textit{overfitting}. This was attributed to the phenomenon where the embedding vectors were dominated by only a few features with very large values, such that most of the features were irrelevant in computing the Euclidean distance between embedding vectors. To force the model to generate more balanced embeddings, we used the $L_2$ regularization described below.
    
\subsection{${L}_2$ Embedding Regularization}
    Realizing that the HBT loss function should provide at least the same results as the standard triplet loss, since it selects the original negative $z_n^b$ or a harder one in the batch, we just needed to overcome the overfitting mentioned in the previous subsection.
    This was done by encouraging Edge2Vec to take advantage of as many features as possible, while avoiding very large values, due to the following $L_2$ regularization:
    \begin{equation}
    \label{eq:l2_reg}
        \mathcal{R}_{L_2} = \sqrt{\frac{1}{3Bd}\sum_{b=1}^{B}{\sum_{i=1}^{d}{(z_a^b[i])^2 + (z_p^b[i])^2 + (z_n^b[i])^2}}}
    \end{equation}
    The final loss function used was thus
    \begin{equation}
        \mathcal{L} = \mathcal{L}_{\text{HBT}} + \lambda\mathcal{R}_{L_2}
    \end{equation}
    where $\lambda = 1$ was found to work best in our case.

\subsection{Intra- and Inter-Image Sampling}
\begin{figure*}
        \centering
        \includegraphics[width=\linewidth]{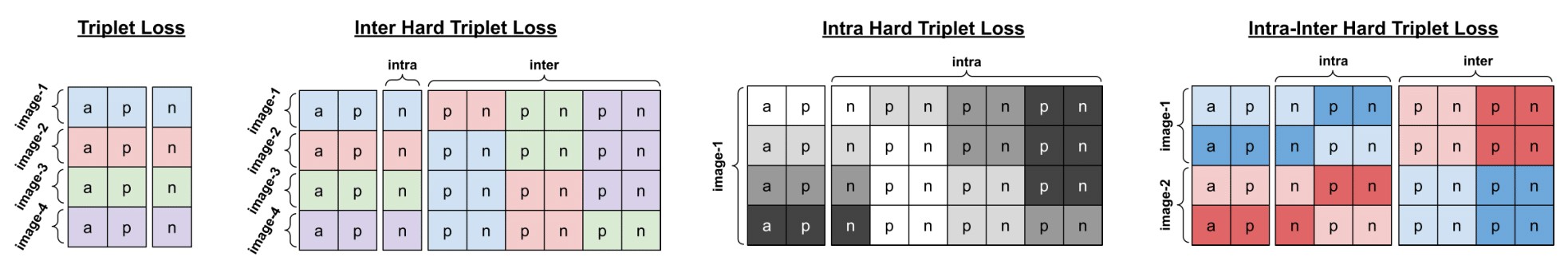}
        \caption{Intra-inter triplet batch with 4 samples. Left to right: Original triplet: every anchor has its own negative; Inter hard batch triplet: each triplet is sampled from a different image in the training set; Intra hard batch triplet: all batch triplets are sampled from same image; Intra-inter hard batch triplet: intra-inter ratio of 1:1}
        \label{fig:intra_inter_sampling}
    \end{figure*}

    The proposed set of HBTs is composed of $B$ triplets, each from a different image (unless $B$ is greater than the number of images in the training set). Thus, the hard negatives per batch sample are \textit{inter} hard negatives, while the only negative from the same image as the anchor ($z_a^b$), is the one from the original triplet, \ie, $z_n^b$. However, it is more likely that distinguishing negatives from other images (\ie, \textit{inter negatives}) will be less difficult than negatives sampled from the anchor's image (\ie, \textit{intra negatives}).
    To maintain a wide range of possibilities, we control the ratio between the numbers of intra and inter samples in a batch, which directly reflects on the ratio between the numbers of intra and inter hard negatives. In principle, we experienced that an intra-inter sampling ratio of 1:1 works the best, since \textit{intra sampling} forces rather discriminative embeddings, while \textit{inter sampling} keeps the training more stable due to a more diverse set of images in the batch.
    Figure~\ref{fig:intra_inter_sampling} illustrates all three combinations of HBTs and also a regular triplet for reference.
    
    Note that using intra sampling increases the chance that some negatives in the batch are the positives of other anchors. Hence, before applying HBT loss to the batch, we need to detect all such instances and remove those potential hard negatives for any relevant anchors in the batch.

\section{Experimental Results}
We have experimented deliberately with the more challenging problem of degraded puzzles containing pieces with \textit{eroded boundaries}, as synthetic/perfect puzzles can be solved rather easily. In particular, while classical CM methods perform very well on the latter type of puzzles, they fail miserably on the former. 

Specifically, we experimented with eroded 1-pixel frame piece boundaries of $28 \times 28$ pieces. Thus, for every piece, we set the outer frame pixels to 0, and retained the original content of the remaining $26 \times 26$ pixels.
The core CNN of our model is identical to TEN's, while the last FC layer was modified according to the \textit{grouping} architecture explained previously.
For our datasets, we used the 800 training images of DIV2K~\cite{Agustsson_2017_CVPR_Workshops} downscaled by a factor of 2 (due to bicubic interpolation), which is the same training set used for TEN. For additional evaluation and to avoid bias as much as possible, we augmented our test set by adding two commonly used datasets. In addition to the 100 images from DIV2K's validation set, we added 100 images from the PIRM dataset~\cite{pirm2018cvpr} and 20 images from the MIT dataset~\cite{conf/cvpr/ChoAF10}, Altogether, we obtained a test set of 220 puzzles, each containing 300--1000 pieces.
For training, we used the Adam optimizer~\cite{kingma2017adam}, a learning rate of 1E-4 (with a decay factor of $0.9$, in case the loss did not decrease for 5 epochs), a batch size of $1024$, and an epoch of $5000$ iterations.
Also, we used Euclidean distance as our distance metric $\mathcal{D}$ and embeddings of size $d=320$.
To ensure symmetric pairwise CMs, we apply post-processing according to Eqs.~\ref{eq:minmax_scaling},~\ref{eq:symmetry}, as was proposed in \cite{ten2022}.
The training was done on a machine with 64 CPU cores, 128GB of RAM and 8 GPUs, each with 11GB memory.
\begin{equation}
\label{eq:minmax_scaling}
    C'(k_i,k_j) = \dfrac{C(k_i,k_j) - min(C(k_i,*))}{max(C(k_i,*)) - min(C(k_i,*))}
\end{equation}
\begin{equation}
\label{eq:symmetry}
    C''(k_i,k_j) = C''(k_j,k_i) = \dfrac{C'(k_i,k_j) + C'(k_j,k_i)}{2}
\end{equation}

\subsection{Grouping: Smaller while Better}
    As noted before, most of the model's parameters (or connection weights) are associated with the final FC layer. Specifically, the number of weights, mapping the spatial CNN features onto a vector $z$ in latent space $\mathbb{R}^d$, depends on the embedding size $d$.
    Since no nonlinear function is applied after the projection layer, each component of $z$ is a weighted sum of all the spatial features.
    We suggest that even if all of the channels in the final layer of spatial features are essential for our embedding $z$, not all of them necessarily contribute to all of $z$'s components.
    Moreover, if not all of the channels should be associated essentially with certain components of $z$, they could actually prove detrimental to the quality of these components.
    Motivated by this insight, we conducted experiments to assess the extent to which the dependencies of each component in $z$ on all channel features $C$ could be down scaled to only a subset of $C / G$ channels.
    Having experimented with $C=512$, $d=120$, and $G \in \{1, 2, 4, 8\}$, we noticed that grouping does not degrade the model's performance; to the contrary, it only resulted in slightly better results (see Table~\ref{tab:ablation}). Moreover, splitting Edge2Vec into $16$ groups decreased the total number of parameters from 9.6M to only 2.1M (\ie, a reduction by a factor of almost 4.6), without any degradation in performance.
    
    \newcolumntype{C}{>{\centering\arraybackslash}X}
    \begin{table}[h]
        \centering
        \caption{Top-1 validation accuracy of all ablation experiments ordered ``chronologically'' from top to bottom; each ablation study contains best hyper parameters found in previous experiments, \eg, if $G$=8 was found for best performance (under \textit{grouping}), all subsequent experiments were conducted with $G$=8.}
        \vspace{0.25cm}
        \label{tab:ablation}
        \begin{tabular}{c}
            % --- Grouping --- %
            \textbf{Grouping} \\
            \begin{tabularx}{0.5\columnwidth}{C C C C}
                 \toprule
                 $G$=1 & $G$=2 & $G$=4 & $G$=8 \\
                 \cmidrule(lr){1-1} \cmidrule(lr){2-2} \cmidrule(lr){3-3} \cmidrule(lr){4-4}
                 65.3\% & 65.7\% & 65.9\% & 66.3\% \\
                %  \bottomrule
            \end{tabularx}
            % --- Embedding Size --- %
            \vspace{6pt}
            \\ \textbf{Embedding Size} \\
            \begin{tabularx}{0.5\columnwidth}{C C C C}
                 \toprule
                 $d$=40 & $d$=80 & $d$=160 & $d$=320 \\
                 \cmidrule(lr){1-1} \cmidrule(lr){2-2} \cmidrule(lr){3-3} \cmidrule(lr){4-4}
                 64.6\% & 66.3\% & 67.1\% & 67.5\% \\
                %  \bottomrule
            \end{tabularx}
            % --- Batch-Size --- %
            \vspace{6pt}
            \\ \textbf{Batch-Size} \\
            \begin{tabularx}{0.5\columnwidth}{C C C C C}
                 \toprule
                 $B$=64 & $B$=128 & $B$=256 & $B$=512 & $B$=1024 \\
                 \cmidrule(lr){1-1} \cmidrule(lr){2-2} \cmidrule(lr){3-3} \cmidrule(lr){4-4} \cmidrule(lr){5-5}
                 66.6\% & 67.1\% & 67.5\% & 68.6\% & 69.6\% \\
                %  \bottomrule
            \end{tabularx}
            % --- Intra-Inter Hard Batch Triplets --- %
            \vspace{6pt}
            \\ \textbf{Intra-Inter Hard Batch Triplets} \\
            \begin{tabularx}{0.5\columnwidth}{C C C}
                 \toprule
                 Inter & Intra-Inter 1:1 & Intra \\
                 \cmidrule(lr){1-1} \cmidrule(lr){2-2} \cmidrule(lr){3-3}
                 69.6\% & 72.9\% & 69.8\% \\
                %  \bottomrule
            \end{tabularx}
        \end{tabular}
    \end{table}

\subsection{HBT Loss and $L_2$ Regularization}
\label{sec:hard_triplet_results}
    As noted before, we should not have observed a worse performance of our model using the HBT loss function (instead of the standard triplet loss), as its basic purpose is to find only harder negatives.
    To explain the overfitting phenomenon encountered initially, we examined the following parameters of embeddings generated by Edge2Vec: (1) The maximal absolute value of an embedding feature, and (2) the fraction of very large embedding components with respect to the embedding size $d$.
    Collecting the various embedding values from the validation set, we found that on average, $22.5\%$ of the absolute values of an embedding's components were in the range $2$--$8$, while all the rest were considerably smaller. And since Euclidean distance is used as the distance metric $\mathcal{D}$, small sets of very large elements could well dominate the outcome of the CM computation. So although HBT loss should force the model to distinguish successfully hard negatives, the learning process has been impaired due to a few dominant sets of values, which affect mostly the gradients' update, while ignoring the remaining $77.5\%$ embedding features containing smaller values.
    Thus, to mitigate the use of very large values in the embedding vector, we added the $L_2$ regularization of  Eq.~\ref{eq:l2_reg} to the final loss function, and controlled its influence by the parameter $\lambda$. Figure~\ref{fig:L2_reg:b} depicts nicely the positive impact of  $L_2$ regularization on the Top-1 accuracy for the test set, which improved from $58.6\%$ to $64.5\%$ for $\lambda=0$ and $\lambda=1$, respectively.
    
    To demonstrate the effect of $L_2$ regularization on the robustness, we gradually masked out the embeddings from the highest absolute value to the lowest, until $50\%$ of an embedding vector were removed. Next, for each level of masking we calculated the Top-1 accuracy on the testset and measured the relative accuracy of the masked embeddings compared to the original unmasked embeddings (which is the original Top-1 accuracy on the testset).
    Figure~\ref{fig:L2_reg:a} depicts the two Top-1 accuracies curves, with and without the $L_2$ regularization. (All of the other training hyperparameters were identical for the two experiments.)
    The above plots imply that Edge2Vec spreads more evenly the embedding features with $L_2$ regularization than without it. (The latter confines most of the embedding energy to only a few features.)
    
    \begin{figure}[h]
        \centering
        \begin{subfigure}{0.4\linewidth}
        \centering
        \includegraphics[width=\linewidth]{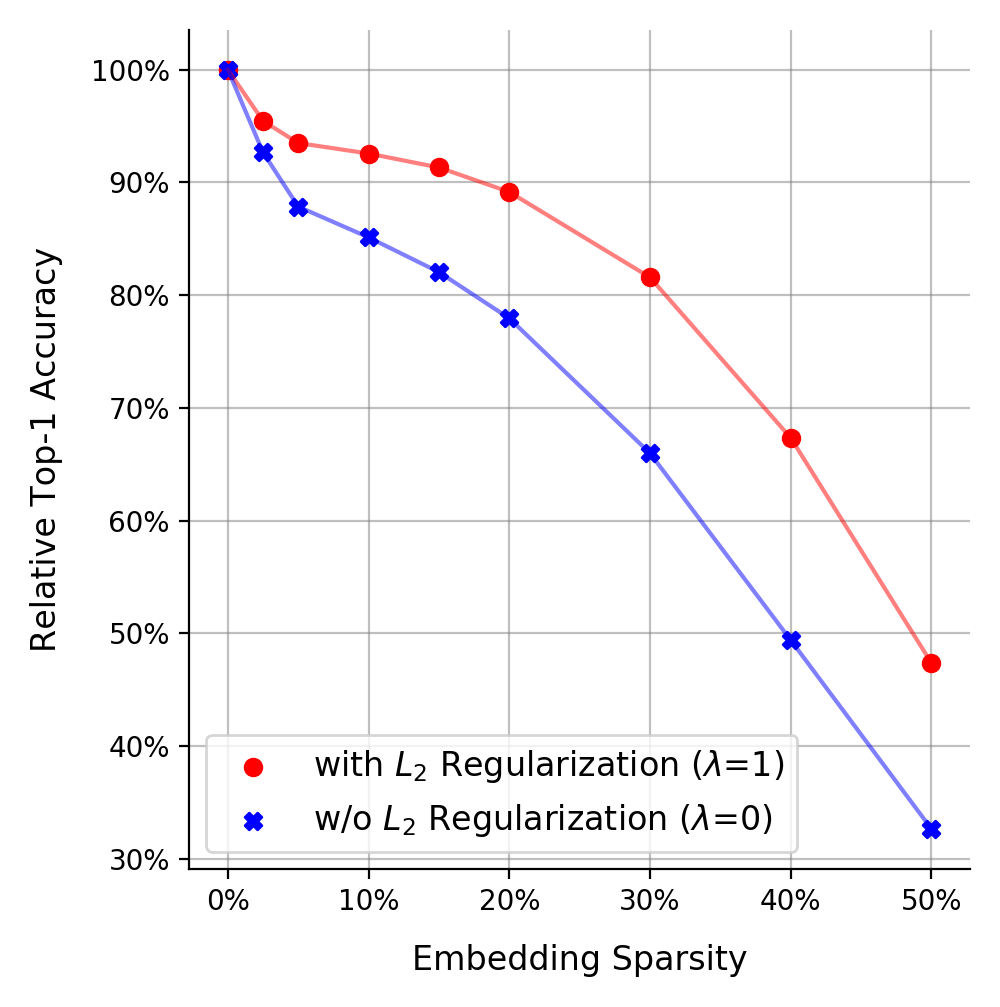}
        \caption{}
        \label{fig:L2_reg:a}
        \end{subfigure}
        % \hfill
        \begin{subfigure}{0.4\linewidth}
        \centering
        \includegraphics[width=\linewidth]{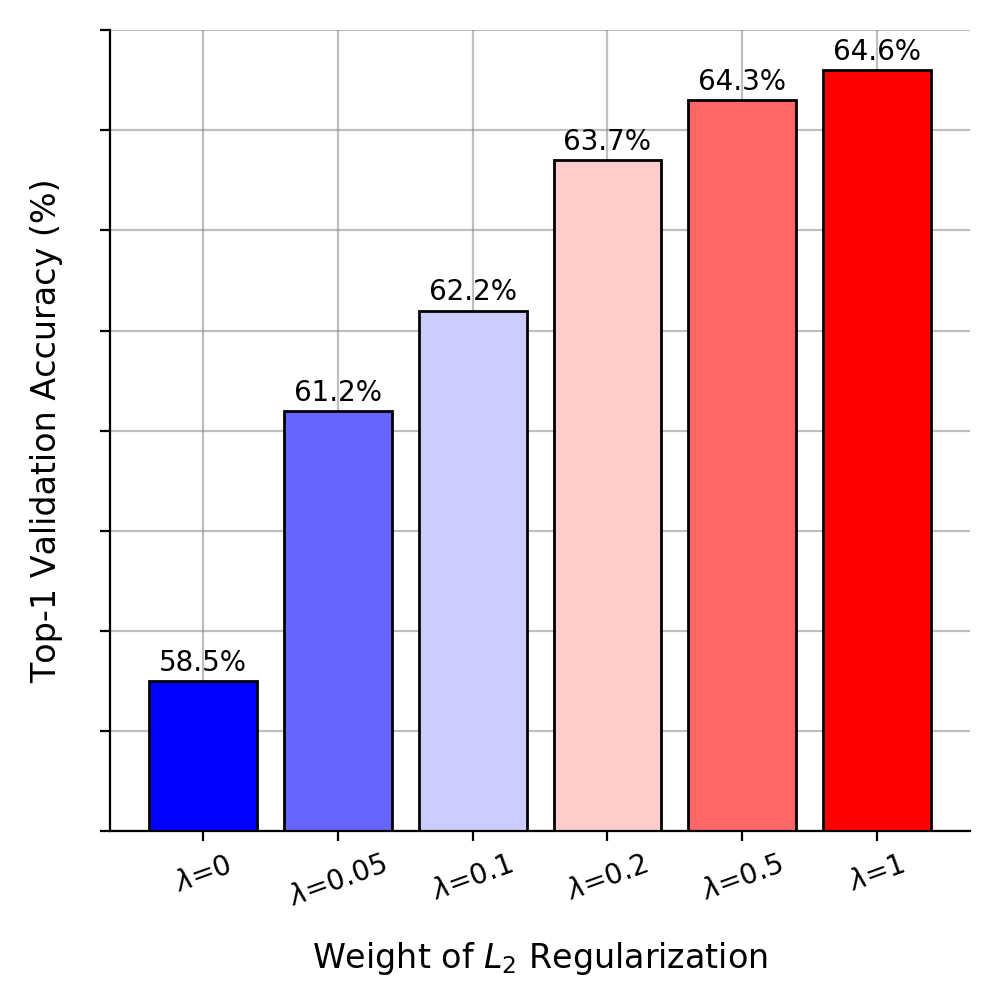}
        \caption{}
        \label{fig:L2_reg:b}
        \end{subfigure}
        \caption{Impact of $L_2$ regularization: (a) Influence of embedding features spread more evenly, \ie, model trained with $L_2$ regularization preserves almost $90\%$ of original Top-1 accuracy with $20\%$ most significant features deleted, whereas model without $L_2$ regularization drops to $\approx 80\%$ of its original Top-1 accuracy with same amount of sparsity; (b) impact of $\lambda$ on final performance: $6\%$ improvement in Top-1 validation accuracy from no regularization ($\lambda=0$) to full regularization ($\lambda=1$).}
        \label{fig:L2_reg}
    \end{figure}
    
\begin{figure}[]
        % \vspace{-0.5cm}
        \centering
            % \begin{subfigure}{0.44\linewidth}
            % \centering
            % \includegraphics[width=\linewidth]{figures/0840x2_24_36_.png}
            % \caption{}
            % \label{fig:the_puzzle:a}
            % \end{subfigure}
            % % \\
            % % \hfill
            % \begin{subfigure}{0.53\linewidth}
            % \centering
            % \includegraphics[width=\linewidth]{figures/distance_map_HBT_2.png}
            % % \includegraphics[width=\linewidth]{figures/distance_map_HBT.png}
            % \caption{}
            % \label{fig:distance_map:b}
            % \end{subfigure}
            
            \begin{subfigure}{0.45\linewidth}
            \centering
            \includegraphics[width=\linewidth]{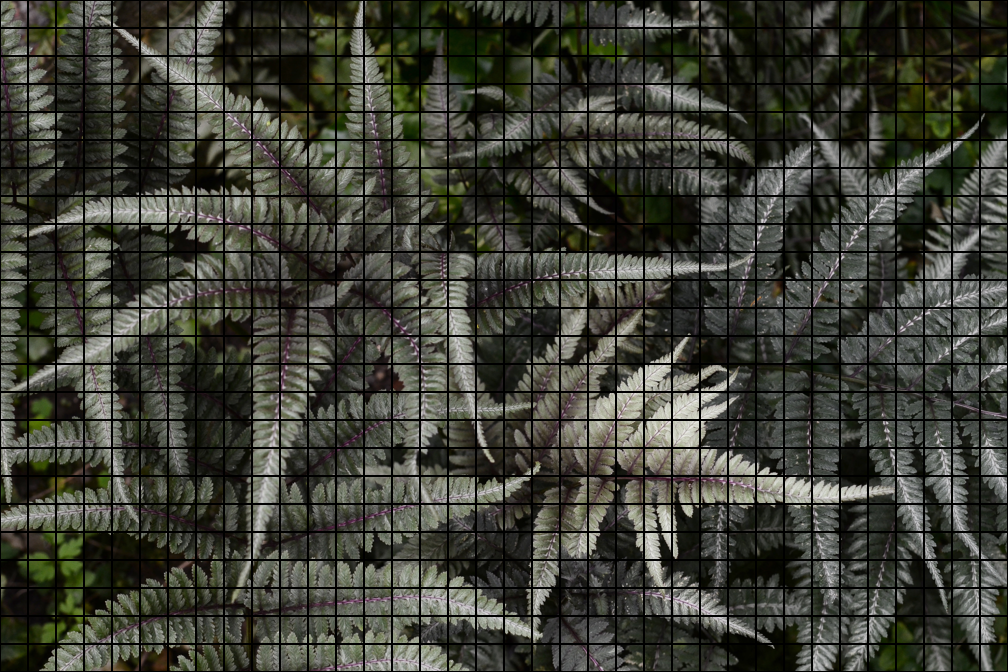}
            \caption{}
            \label{fig:the_puzzle:a}
            \end{subfigure}
            % \\
            \hfill
            \begin{subfigure}{0.53\linewidth}
            \centering
            \includegraphics[width=\linewidth]{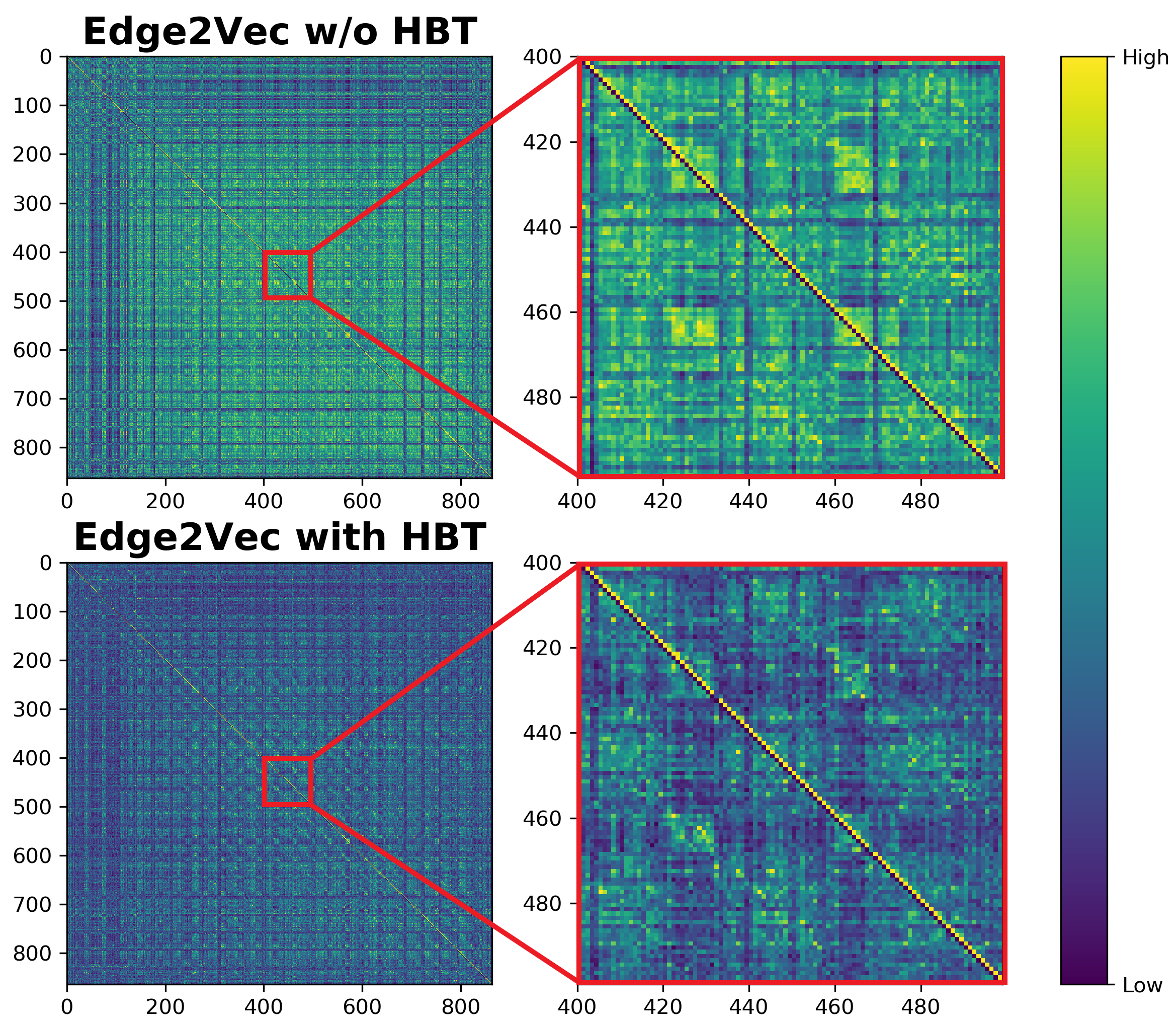}
            \caption{}
            \label{fig:distance_map:b}
            \end{subfigure}
            
        \caption{Distance map visualization: (a) 864-piece puzzle from DIV2K testset; (b) distance map of Edge2Vec without (top)/with (bottom) HBT loss; rows represent an anchor piece $N$, and columns represent the examined piece with respect to \textit{right} edge of the anchor; unless anchor is located at the rightmost column of puzzle, its correct neighboring piece on the \textit{right} is $N+1$. Namely,  compatibility scores along super diagonal of CM matrix should be largest (in yellow), while the rest should be (much) smaller (purple); $100 \times 100$ zoom-in maps clearly demonstrate enhanced discrimination with HBT loss.}
        \label{fig:distance_map}
    % \vspace{-0.5cm}
    \end{figure}

    Figure~\ref{fig:distance_map} demonstrates the combined effect of our HBT loss and $L_2$ regularization on an 864-piece puzzle from the DIV2K dataset. Specifically, it demonstrates the dramatic improvement, due to the use of HBT, of the discrimination (in terms of CM scores) between correct neighboring pieces (shown in yellow on the matrix super diagonal, with respect to corresponding anchors on the main diagonal) versus all other candidates (shown in dark).
    %We picked a 864-pieces puzzle from the DIV2K test-set, and plot the distance map of our Edge2Vec with and without the proposed HBT loss with $L_2$ regularization.
    %As can be seen in Figure~\ref{fig:distance_map}, using the HBT loss improved dramatically the discrimination between the correct pieces (the 1-pixel upper diagonal) to all other candidates.
    
% \subsection{Impact of Color Channel Augmentation}
% \label{sec:impact_of_color_channel_augmentation}

\subsection{Dimensionality and Batch Size}
\label{dimensionality_and_batchsize}
    \textbf{Dimensionality.} The core of this work is to represent a piece with respect to one of its edges by a vector $z$ in the latent space $\mathbb{R}^d$. The
    vector dimensionality $d$ is the number of features that Edge2Vec works with, and could accidentally become a performance bottleneck when setting it too low.
    Thus, during training we experimented with various $d$ values, \ie, $d \in \{ 40, 80, 160, 320 \}$. As indicated by Table~\ref{tab:ablation}, it is desirable to set $d \ge 160$; the best performance (through our experiments) was obtained for $d=320$.
    
    \textbf{Batch-size.} It is well known that a larger batch size could assist in the training, as the gradients become more stable.
    In our case, the batch size highly affects our proposed HBT loss function, since more samples in the batch can potentially generate harder negatives as explained before.
    Hence, we tested Edge2Vec on a variety of batch sizes as shown in Table~\ref{tab:ablation}.
    The results reveal that typically a larger batch size causes Edge2Vec to extract more discriminative features, which results in better performance.
    
\subsection{Performance Evaluation}
\label{performance_evaluation}
\begin{figure}[h]
        \centering
        \includegraphics[width=0.7\linewidth]{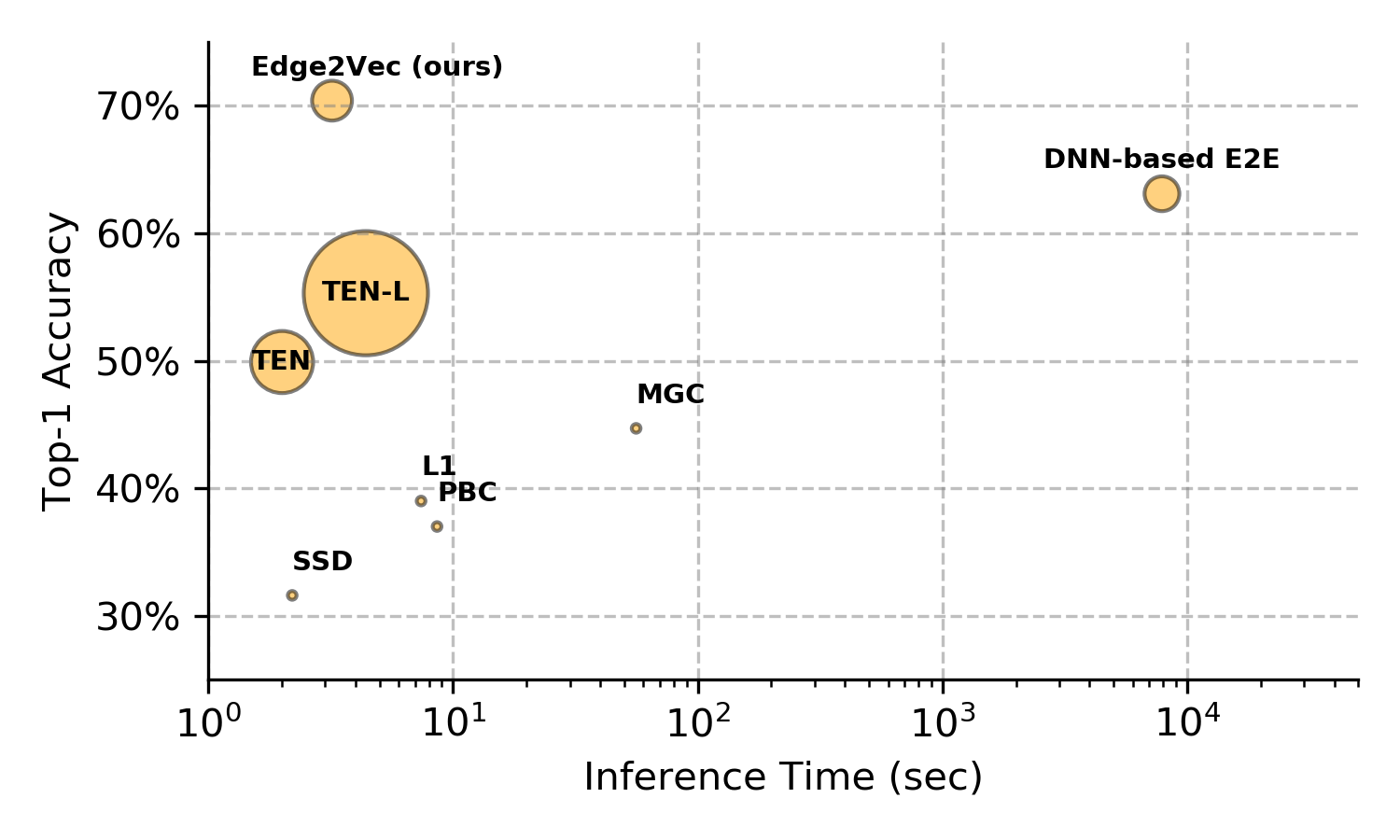}
        \caption{Trade-off illustration for all evaluated CMs on puzzles with \textit{eroded boundaries}, regarding: (1) Average Top-1 accuracy ($y$-axis) for Type-2 on the 3 testsets; (2) model footprint (depicted by bubble size); and (3) inference (wall-clock) time ($x$-axis) for computing all $16N(N-1)$ scores of 1,600-piece puzzle;
        %as roughly half of the CMs evaluated are DNN-based, it is of interest to note also the number of parameters of each model; 
        although TEN-like model has much larger footprint than E2E (because of larger last fully-connected layer and use of two identical models (rather than one)), it is much faster because of $8N$ executions (instead of $16N^2$); tiny bubbles depict classical CM schemes, which lack footprint. Our Edge2Vec CM provides best trade-off wrt above criteria. }
        \label{fig:bubbles}
    \end{figure}

    After picking the best hyperparameters for our Edge2Vec model, according to the empirical ablation study reported, we ran a comparative performance evaluation against previous CM models.
    Before analyzing, in more detail, our experimental evaluation, we note that a CM model could be characterized by the following three main attributes: (1) Top-1 accuracy, (2) \textit{footprint} and computational complexity (relevant to all DNN-based methods), and (3) actual inference time. Figure~\ref{fig:bubbles} below clearly indicates that the best trade-off with respect to the above criteria is achieved by Edge2Vec, compared to all of the classical and DNN-based CMs evaluated.

    \subsubsection{Top-1 Accuracy}
    \label{top_1_accuracy}
        Recall, for an $N$-piece puzzle, a piece edge has $N-1$ or $4(N-1)$ matching candidates for Type-1 or Type-2, respectively.
        We say that a piece edge is Top-1 if its most compatible edge candidate (according to a CM model in question) is its very adjacent edge in the original puzzle, \ie, its ``ground truth'' neighboring edge.
        Following the above Top-1 definition, Top-1 accuracy is the percentage of piece edges that are Top-1 according to a specific CM algorithm. (For a set of multiple puzzles, we can regard Top-1 accuracy as the average Top-1 accuracy over all puzzles in the dataset.) The higher the Top-1 accuracy for a given CM module, the better its performance. Ideally, Top-1 accuracy should approach $100\%$.
        
        We compared our Edge2Vec to various classical and DNN-based E2E CMs, and found that our method achieved higher Top-1 accuracies than those obtained by the previous embedding-based TEN-L (by 12.2\% and 15.5\% for Type-1 and Type-2, respectively). Moreover, it also outperformed the highly-intensive DNN-based E2E method, as summarized in Table~\ref{tab:performance_eval}.
        
\subsubsection{Footprint and Computational Complexity}
    \label{footprint}
    The notion of \textit{zero emission} has been gaining recently more attention, as to the operation volume of NN-based models, in terms of footprint and computational complexity.
    %is another important aspect to be taken into consideration when training and working with NN-based models.
    Thus, we compared naturally Edge2Vec to previous NN-based CMs also with respect to these terms. Specifically, footprint is measured by a model's number of parameters, and computational complexity is measured by the number of \textit{multiply–accumulates} (MACs) per a single pair of pieces, or per an entire $N$-piece puzzle (for a total of $16N^2$ pairs). The number of parameters and the number of MACs per single pair are provided typically by a library in use, whereas the number of MACs for an  $N$-piece puzzle is implied by the nature of a model in question (see below). 
    
    \textbf{Number of MACs for CM via DNN-based E2E:} Given a pair of pieces, each of size $28 \times 28$ pixels, the input size of a DNN-based E2E CM is (28, 56, 3), and the number of MACs is just the number of multiplications and additions during a forward pass.
    Thus, the CM complexity of a DNN-based E2E for an $N$-piece puzzle is the complexity per a single pair multiplied by $16N^2$ (\ie, multiplied by the number of pair combinations).
    
    \textbf{Number of MACs for CM via DNN-based with Embeddings:} To compute, in this case, the CM of a single pair of pieces $(k_{right}, k_{left})$, we first represent them in latent space, due to some NN-based model (\eg, TEN, TEN-L or Edge2Vec), obtaining $(z_{right}, z_{left})$. We then calculate the Euclidean distance between these embeddings. Since the execution of NN-based models for each pair piece is much more intensive than the computation of the above Euclidean distance, we consider, essentially, only the number of MACs involving the projection of the two pieces to embedding space via an embedding-based model in question. In other words, this should be multiplied, essentially, just by the total number of embeddings, to obtain the (approximate) number of MACs for an entire $N$-piece puzzle.      
    %In contrast to CM computations via DNN-based E2E, calculating all $16N^2$ pairwise permutations (for an $N$-piece puzzle) isn't a simple multiplication of the complexity per a single pair.
    Exploiting the notion of our embedding-based models, we need only consider, therefore,  
    $4N$ embeddings with respect to all (potential) boundaries of each piece, \ie, multiply by a factor of $8N$ embedding operations (as each piece embedding needs to be considered twice, for the piece itself and for its flipped version).    
    %the embeddings of all the $4N$ boundaries wrt right and left directions, resulting in a total of $8N$ executions of the NN-based embedding model.
    %Next, all we need to do, is to execute $16N^2$ times an Euclidean distance between pairs of embedding vectors.
    See Table~\ref{tab:footprint_and_complexity} for specific results. (As noted, the number of MACs neglects the $16N^2$ Euclidean distance operations, as their execution is considerably faster than the embedding computations themselves.)
    %only represents the first stage, since the latter phase has less complexity in a few orders of magnitude.
    
    \newcolumntype{A}{>{\centering}p{4cm}}
    \newcolumntype{B}{>{\centering}p{2cm}}
    \newcolumntype{C}{>{\centering\arraybackslash}p{3.5cm}}
    \begin{table*}[h]
        \centering
        \caption{Footprint and computational complexity. Our Edge2Vec CM has a competitive footprint and extremely low complexity; note the linear dependence of \# MACs wrt $N$ for embedding-based models, in contrast to its quadratic dependence wrt $N$ for DNN-based E2E; although TEN-L is substantially bigger than other NN-based CMs (being an ensemble of TEN), it remains very efficient wrt DNN-based E2E CMs, for the entire puzzle, due to embedding.}
        %The most important part is the number of MACs(G) when calculating the CM for an entire $N$-piece puzzle - TEN, TEN-L and Edge2Vec have linear complexity wrt $N$ while DNN-based E2E has a quadratic complexity wrt $N$.}
        \vspace{0.25cm}
        \begin{tabular}{ l B c B C }
            % \topruled
            \multirow{2}{*}{\textbf{Model}} & \multirow{2}{*}{\textbf{\# Params (M)}} && \multicolumn{2}{c}{\textbf{\# MACs (G)}} \\
            \cmidrule(l){4-5}
             &  && \textbf{Single pair} & \textbf{$N$-piece puzzle} \\
            \midrule
            \textbf{TEN} & 5.1 && 0.35 & 1.4$N$ \\
            \textbf{TEN-L} & 20.4 && 1.4 & 5.6$N$ \\
            \textbf{DNN-based E2E} & 1.6 && 0.35 & 5.6$N^2$ \\
            \textbf{Edge2Vec (ours)} & 2.1 && 0.35 & 1.4$N$ \\
            \bottomrule
        \end{tabular}
        \label{tab:footprint_and_complexity}
    \end{table*}

\subsubsection{Inference Time}
    \label{efficiency}
    Edge2Vec retains, essentially, TEN's faster running times, due to our novel use of embeddings as an intermediate step in the CM computation pipeline. See Table~\ref{tab:inference_times} for wall-clock times of all NN-based CMs, which compute all $16N^2$ compatibility scores for different puzzle sizes. The running time of Edge2Vec is between that of TEN and TEN-L, \ie, much faster than that of the intensive DNN-based E2E architecture. (TEN's slightly faster inference times could be attributed to its smaller embedding dimension, \ie, 40, instead of 320 used by Edge2Vec.)
    
    \newcolumntype{A}{>{\centering}p{2cm}}
    \newcolumntype{D}{>{\centering}p{1.5cm}}
    \newcolumntype{B}{>{\centering}p{3cm}}
    \newcolumntype{C}{>{\centering\arraybackslash}p{3cm}}
    \begin{table*}[h]
        \centering
        \caption{Wall-clock inference times (\textbf{in seconds}) of our proposed Edge2Vec, compared to previous embedding-based CMs and DNN-based E2E; Edge2Vec remains extremely efficient compared to E2E mode.}
        \vspace{0.25cm}
        \begin{tabular}{ D B B B C }
            % \toprule
            % \textbf{\# Pieces} & \textbf{TEN \scriptsize{[sec]}} & \textbf{Edge2Vec (ours) \scriptsize{[sec]}} & \textbf{TEN-Large \scriptsize{[sec]}} & \textbf{DNN-based E2E \scriptsize{[sec]}} \\
            \textbf{\# Pieces} & \textbf{TEN} & \textbf{Edge2Vec (ours)} & \textbf{TEN-L} & \textbf{DNN-based E2E} \\
            \midrule
            \textbf{100} & $7.0 \times 10^{-2}$ & $1.4 \times 10^{-1}$ & $1.8 \times 10^{-1}$ & $3.2 \times 10^{1}$ \\
            % \hline
            \textbf{200} & $1.4 \times 10^{-1}$ & $2.6 \times 10^{-1}$  & $3.9 \times 10^{-1}$ & $1.2 \times 10^{2}$ \\
            % \hline
            \textbf{400} & $3.6 \times 10^{-1}$ & $5.2 \times 10^{-1}$  & $8.9 \times 10^{-1}$ & $4.9 \times 10^{2}$ \\
            % \hline
            \textbf{800} & $1.0 \times 10^{0}$ & $1.5 \times 10^{0}$  & $2.2 \times 10^{0}$ & $1.9 \times 10^{3}$ \\
            % \hline
            \textbf{1600} & $3.2 \times 10^{0}$ & $4.2 \times 10^{0}$  & $6.2 \times 10^{0}$ & $7.9 \times 10^{3}$ \\
            % \hline
            \textbf{3200} & $1.1 \times 10^{1}$ & $1.4 \times 10^{1}$  & $1.9 \times 10^{1}$ & $3.1 \times 10^{4}$ \\
            \bottomrule
        \end{tabular}
        \label{tab:inference_times}
    \end{table*}
        
       % We compared our Edge2Vec to various classical and DNN-based E2E CMs, and found that our method achieved higher Top-1 accuracies than those obtained by the previous embedding-based TEN-L (by 12.2\% and 15.5\% for Type-1 and Type-2, respectively). Moreover, it also outperformed the highly-intensive DNN-based E2E method, as summarized in Table~\ref{tab:performance_eval}. 
    
    \subsubsection{Reconstruction Results}
    \label{reconstraction_results}
        Although Top-1 accuracy is a useful metric for evaluating the quality of a CM algorithm, we recall that CM's functional role is to assist in the eventual reconstruction of a given puzzle, as accurately as possible.
        Hence, in addition to Top-1 accuracy, we also conducted a reconstruction comparison between all of the CMs evaluated.
        For completeness, we chose two reconstruction algorithms: Gallagher's reconstruction algorithm~\cite{gallagher2012jigsaw} and a genetic algorithm (GA)-based solver~\cite{rika2019gecco}.
        We decided to work with these two methods, since Gallagher's solver is publicly available and commonly used in many other jigsaw puzzle studies. However, since Gallagher's solver is a greedy algorithm, which could easily get trapped in local optima, we also tested the CMs in conjunction with the above GA-based solver, which is a global optimization method.
        
        In employing Gallagher's greedy reconstruction algorithm to supplement any of the evaluated CMs, we used his rescaling technique for the pairwise CMs
        \begin{equation}
        \label{eq:gallagher_processing}
            \hat{C}(k_i, k_j) = \frac{C(k_i, k_j)}{Second(C(k_i, *))}
        \end{equation}
        where $Second(C(k_i, *))$ denotes the second most compatible value for anchor $k_i$, before introducing their scores into the reconstruction phase.
        The goal of this operation, is to give a degree of confidence to the preliminary pairwise CM values, which can help a lot in a greedy reconstruction algorithm like Gallagher's.
        Table~\ref{tab:performance_eval} contains all of the reconstruction results, measured by the \textit{neighbor accuracy} criterion, for Type-1 and Type-2, using Gallagher's algorithm and the GA-based solver. It can be seen that our proposed Edge2Vec model improved the reconstruction results using DNN-based E2E with both reconstruction algorithms.
        Specifically, using Edge2Vec with Gallagher's solver achieved an average improvement across all three test sets of 5.8\% and 19.5\% for Type-1 and Type-2, respectively. Using Edge2Vec in conjunction with the GA-based solver improved the average reconstruction rates of Type-1 and Type-2 by 1.8\% and 8.3\%, respectively.

        \begin{table*}
        \centering
        
        \caption{Comparative evaluation of various CMs for Type-1 (top) and Type-2 (bottom), with respect to Top-1 accuracy (left) and neighbor accuracy due to Gallagher's reconstruction (middle) and GA-based solver (right); average accuracy (over 10 runs for each puzzle) reported for latter reconstruction. Edge2Vec exhibits superior performance to all other methods (in nearly all cases); best results appear in bold.}
        
        \begin{tabular}{ l c c c  c  c c c  c  c c c }
        
        \multicolumn{12}{c}{\textbf{TYPE-1}} \\
        
        \toprule
        
        \multirow{2}{*}{\textbf{Method}} & \multicolumn{3}{c}{\textbf{Top-1 Accuracy}} && \multicolumn{3}{c}{\textbf{Gallagher's solver}} && \multicolumn{3}{c}{\textbf{GA-based solver}} \\
        \cmidrule(l){2-4} \cmidrule(l){6-8} \cmidrule(l){10-12}
        
        & \textbf{DIV2K} & \textbf{PIRM} & \textbf{MIT} && \textbf{DIV2K} & \textbf{PIRM} & \textbf{MIT} && \textbf{DIV2K} & \textbf{PIRM} & \textbf{MIT} \\
        \midrule
        
        SSD & 37.4\% & 41\% & 42\% && 28.7\% & 32.4\% & 34.5\% && 27.9\% & 34\% & 32.4\% \\
        PBC & 46.4\% & 49.1\% & 43.7\% && 43.7\% & 47.1\% & 36.4\% && 49.3\% & 53.2\% & 44.7\% \\
        $L_1$ & 47\% & 49.8\% & 46.6\% && 42.2\% & 46\% & 41.3\% && 43.6\% & 50.9\% & 43.2\% \\
        MGC & 56.3\% & 59.5\% & 53.5\% && 54.7\% & 60.2\% & 49.4\% && 64.4\% & 71.4\% & 58.2\% \\
        TEN & 59.4\% & 61\% & 55.2\% && 52.9\% & 53.1\% & 44.9\% && 67.5\% & 71.5\% & 61.2\% \\
        TEN-L & 64.6\% & 65.4\% & 59.1\% && 60.7\% & 59.8\% & 53.4\% && 72.9\% & 76.8\% & 65.1\% \\
        DNN-based E2E & 71.4\% & 72.6\% & 69.4\% && 75.4\% & 76.1\% & 69\% && 82.5\% & 90.8\% & \textbf{87.3\%} \\
        \midrule
        Edge2Vec & \textbf{76.4\%} & \textbf{77.8\%} & \textbf{71.3\%} && \textbf{80.2\%} & \textbf{82.3\%} & \textbf{75.3\%} && \textbf{87.4\%} & \textbf{92.1\%} & 86.5\% \\
        \bottomrule
        \\
        \multicolumn{12}{c}{\textbf{TYPE-2}} \\
        \toprule
        \multirow{2}{*}{\textbf{Method}} & \multicolumn{3}{c}{\textbf{Top-1 Accuracy}} && \multicolumn{3}{c}{\textbf{Gallagher's solver}} && \multicolumn{3}{c}{\textbf{GA-based solver}} \\
        \cmidrule(l){2-4} \cmidrule(l){6-8} \cmidrule(l){10-12}
        
        & \textbf{DIV2K} & \textbf{PIRM} & \textbf{MIT} && \textbf{DIV2K} & \textbf{PIRM} & \textbf{MIT} && \textbf{DIV2K} & \textbf{PIRM} & \textbf{MIT} \\
        \midrule
        SSD & 29\% & 32.2\% & 33.5\% && 7.3\% & 9.8\% & 10.3\% && 16.3\% & 18.4\% & 18.3\% \\
        PBC & 37.5\% & 39.3\% & 34.1\% && 13.8\% & 16.5\% & 9.5\% && 26.1\% & 30.5\% & 20.9\% \\
        $L_1$ & 38.7\% & 40.7\% & 37.6\% && 12.2\% & 15\% & 10.7\% && 25.6\% & 28.2\% & 22\% \\
        MGC & 45.4\% & 47.5\% & 41.1\% && 19.7\% & 22.7\% & 12.8\% && 40.2\% & 47.5\% & 32.3\% \\
        TEN & 50.5\% & 52.5\% & 46.6\% && 21.8\% & 23.4\% & 16.6\% && 46.9\% & 50.9\% & 39.6\% \\
        TEN-L & 56.8\% & 57.5\% & 51.5\% && 30.2\% & 33.6\% & 22.7\% && 55.5\% & 59.4\% & 49.1\% \\
        DNN-based E2E & 64.1\% & 64.2\% & 60.9\% && 48.5\% & 49.9\% & 39\% && 73.3\% & 79.7\% & 74.5\% \\
        \midrule
        Edge2Vec & \textbf{71.4\%} & \textbf{74.2\%} & \textbf{66.9\%} && \textbf{65.2\%} & \textbf{70.7\%} & \textbf{60\%} && \textbf{82.3\%} & \textbf{89\%} & \textbf{81.2\%} \\
        \bottomrule
        \end{tabular}
        \label{tab:performance_eval}
        \end{table*}
    
    \subsubsection{Comparison to State-of-the-Art Model}
    \label{comparison_to_sota_model}
        for the completeness of the experiments, we compared Edge2Vec to the previous SOTA method by Bridger et al~\cite{bridger2020cvpr}. As their model requires $64 \times 64$ pieces, we trained our Edge2Vec on the same piece size with similar erosion (1 eroded pixel of a $28 \times 28$ piece is equivalent to about 2 eroded pixels of a $64 \times 64$ piece). We then used Gallagher’s open source solver~\cite{gallagher2012jigsaw} for comparative evaluation, due to a lack of code availability of Paikin and Tal’s solver~\cite{paikin2015solving}. In any case, ~\cite{gallagher2012jigsaw}’s solver should be inferior compared to the latter, so our relative results could only improve.
        We also compared our wall-clock run-times per puzzle size against those provided by \cite{bridger2020cvpr}. (for CM computation and reconstruction).
        See at Table~\ref{tab:sota_comparison} a reconstruction comparison and inference time between the results in \cite{bridger2020cvpr} and ours.
        In summary, our Edge2Vec is accurate and also way more efficient than the method proposed by \cite{bridger2020cvpr}.

\section{Conclusions}
In this paper we proposed a new CM scheme called Edge2Vec. Our newly derived CM uses embeddings in an efficient yet highly-accurate manner. In particular, we demonstrated how the use of a \textit{single embedding model} combined with our newly proposed \textit{hard batch triplet loss} function results in superior performance relatively to that of previous traditional CMs, as well as previous DNN-based E2E CMs, in terms of both Top-1 accuracy and reconstruction neighbor accuracy. To the best of our knowledge, Edge2Vec currently exhibits the best trade-off between precision, efficiency, and footprint.

\newpage

\begin{table}[]
\vspace{-2cm}
    \centering
    \caption{Comparison to state-of-the-art Model. As can be seen, our Edge2Vec CM combined with Gallagher's solver~\cite{gallagher2012jigsaw} was able to achieve superior performance in all metrics compared to the previously state-of-the-art method~\cite{bridger2020cvpr}, which is a GAN-based E2E CM combined with Paikin and Tal's solver~\cite{paikin2015solving}. It is worth to mentioned that our proposed Edge2Vec was able to produced these results with less powerful reconstruction algorithm, while better can be done by only using other reconstruction algorithm.}
    \vspace{0.3cm}
    
    \begin{tabular}{ l  c c  c  c c  c  c c }
    
    \toprule
    
    \multirow{2}{*}{\textbf{Dataset}} & \multicolumn{2}{c}{\textbf{Neighbor accuracy (\%)}} && \multicolumn{2}{c}{\textbf{Perfect reconstruction}} && \multicolumn{2}{c}{\textbf{Running time (sec)}} \\
    \cmidrule(l){2-3} \cmidrule(l){5-6} \cmidrule(l){8-9}
    
    & \textbf{\cite{bridger2020cvpr}} & \textbf{Edge2Vec} && \textbf{\cite{bridger2020cvpr}} & \textbf{Edge2Vec} && \textbf{\cite{bridger2020cvpr}} & \textbf{Edge2Vec} \\
    \midrule
    MIT (70 pieces) & 84.6 & 92.4 (+7.8) && 4 & 10 (+6) && 62 (x149) & 0.415 \\
    McGill (88 pieces) & 76.9 & 83.7 (+6.8) && 7 & 7 (+0) && 98 (x196) & 0.498 \\
    BGU (150 pieces) & 76.3 & 91.5 (+15.2) && 2 & 5 (+3) && 286 (x293) & 0.976 \\
    \bottomrule
    \end{tabular}
    \label{tab:sota_comparison}
\end{table}

\begin{figure}[]
% \vspace{1cm}
\centering
    \begin{tabular}{c c c c}
    
        \includegraphics[width=0.2111\linewidth]{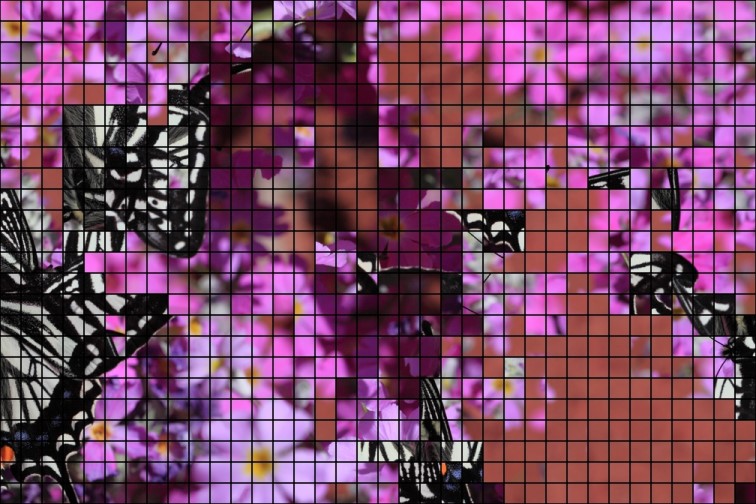} &
        \includegraphics[width=0.2111\linewidth]{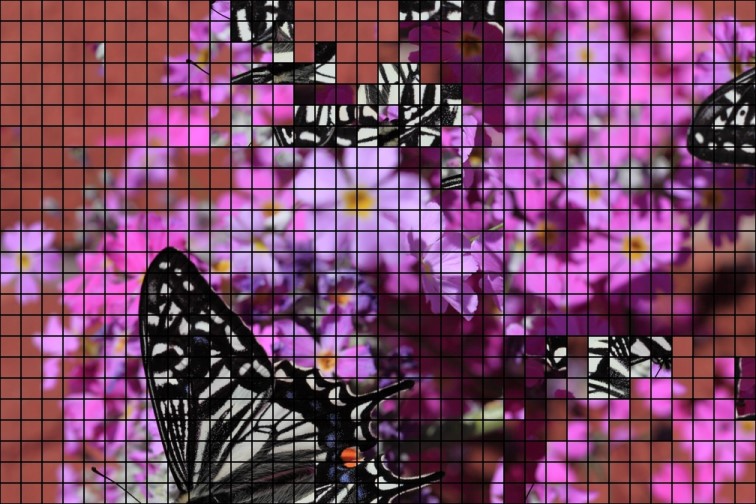} &
        \includegraphics[width=0.2111\linewidth]{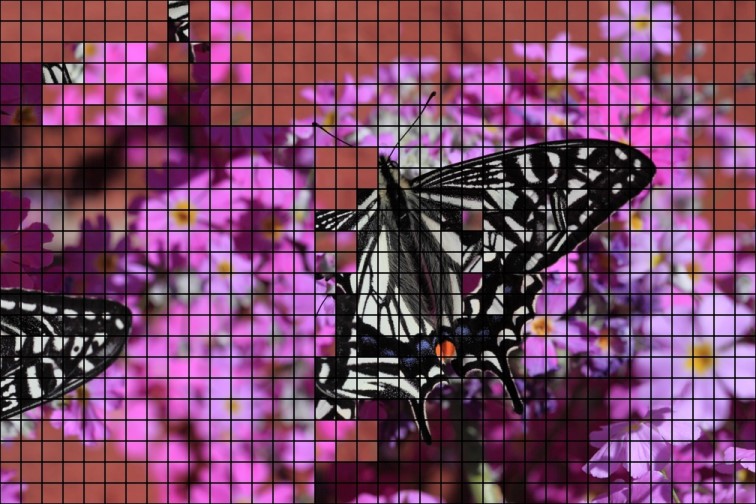} &
        \includegraphics[width=0.2111\linewidth]{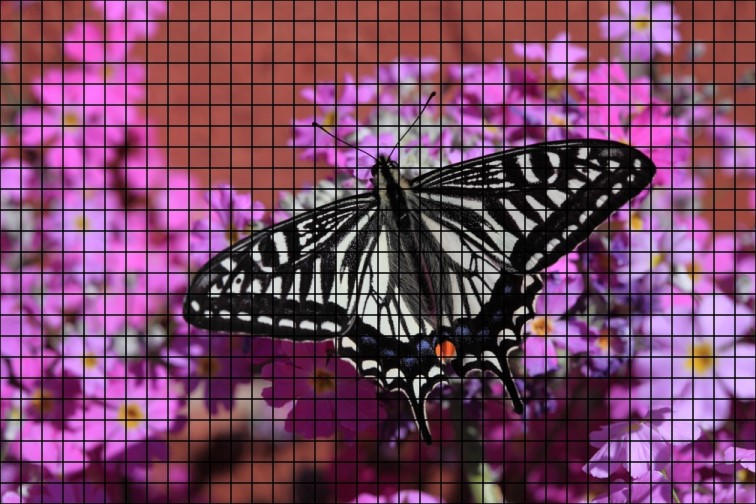}
        \\ \footnotesize{TEN 17.6\%} & \footnotesize{TEN-L 46.7\%} & \footnotesize{DNN-Based E2E 48.7\%} & \footnotesize{Edge2Vec 93.7\%} \\
        
        \includegraphics[width=0.2111\linewidth]{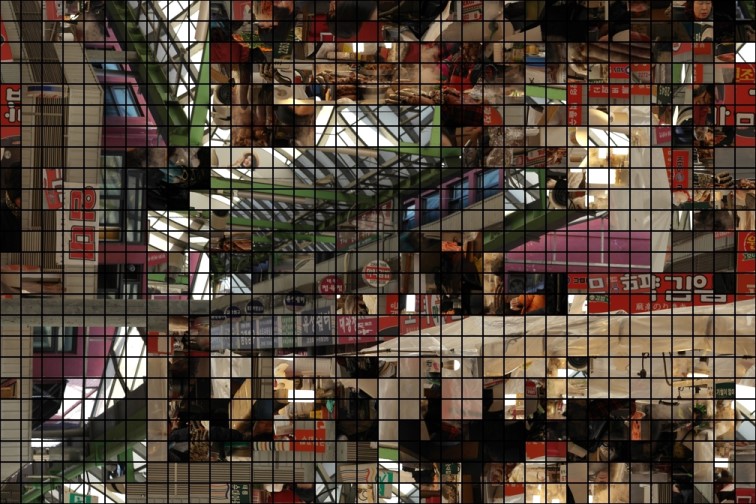} &
        \includegraphics[width=0.2111\linewidth]{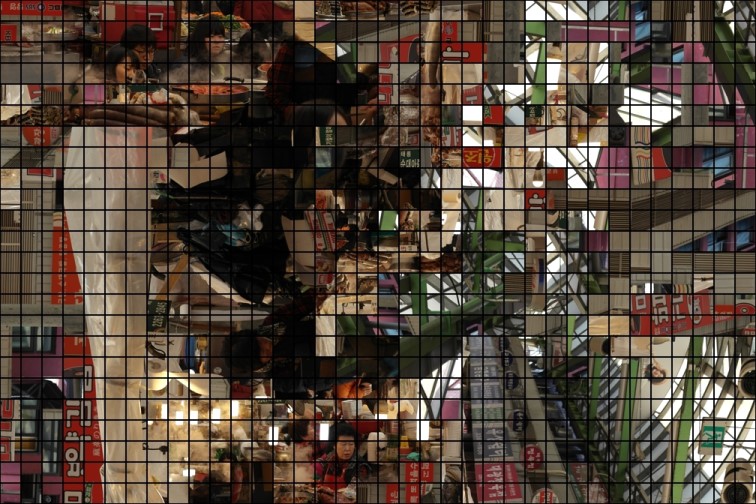} &
        \includegraphics[width=0.2111\linewidth]{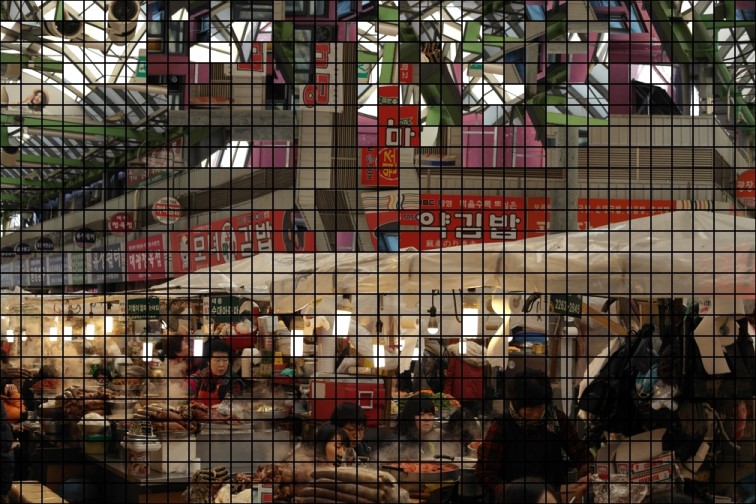} &
        \includegraphics[width=0.2111\linewidth]{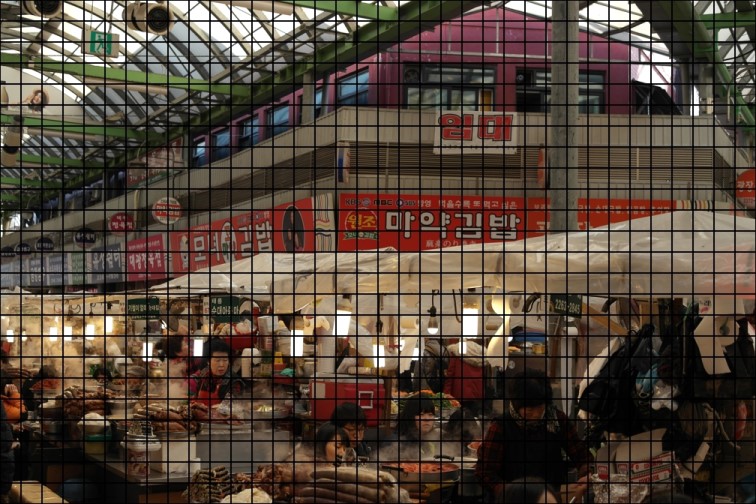}
        \\ \footnotesize{TEN 19.9\%} & \footnotesize{TEN-L 21.8\%} & \footnotesize{DNN-Based E2E 72\%} & \footnotesize{Edge2Vec 100\%} \\

        % \toprule \\
        
        \includegraphics[width=0.2111\linewidth]{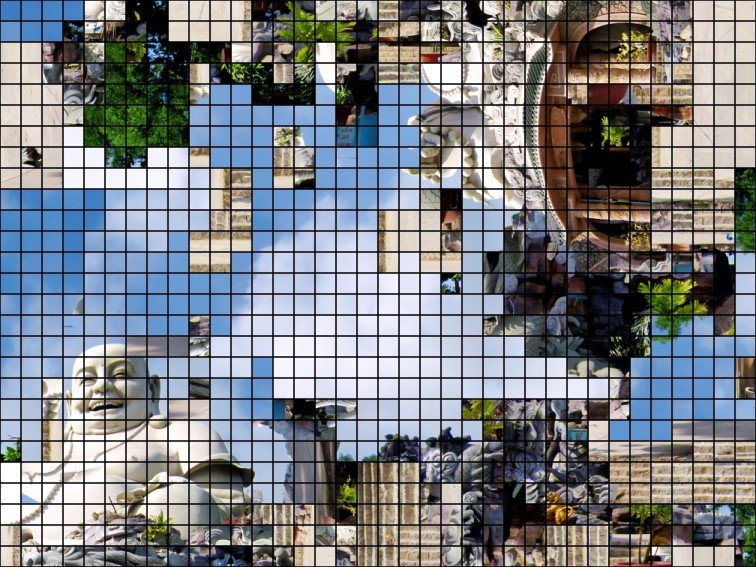} &
        \includegraphics[width=0.2111\linewidth]{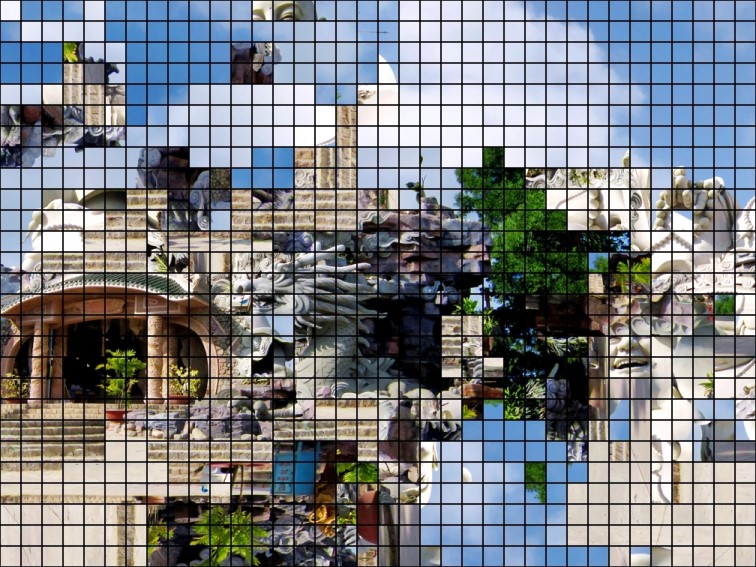} &
        \includegraphics[width=0.2111\linewidth]{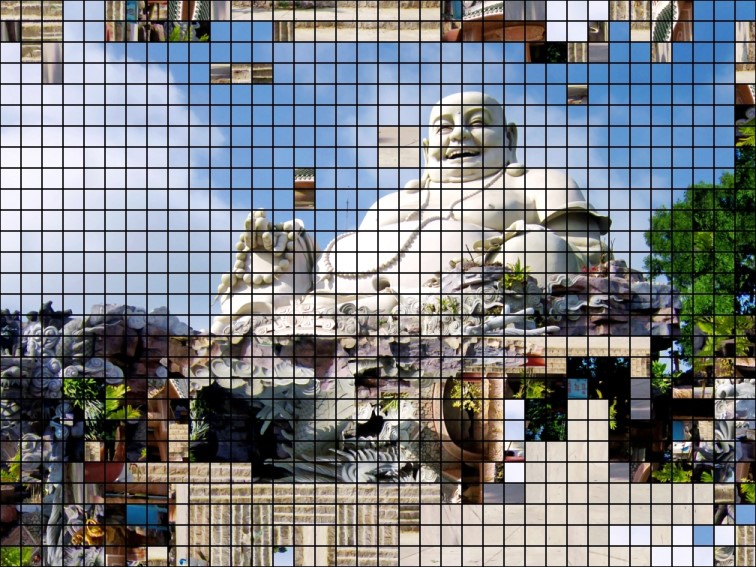} &
        \includegraphics[width=0.2111\linewidth]{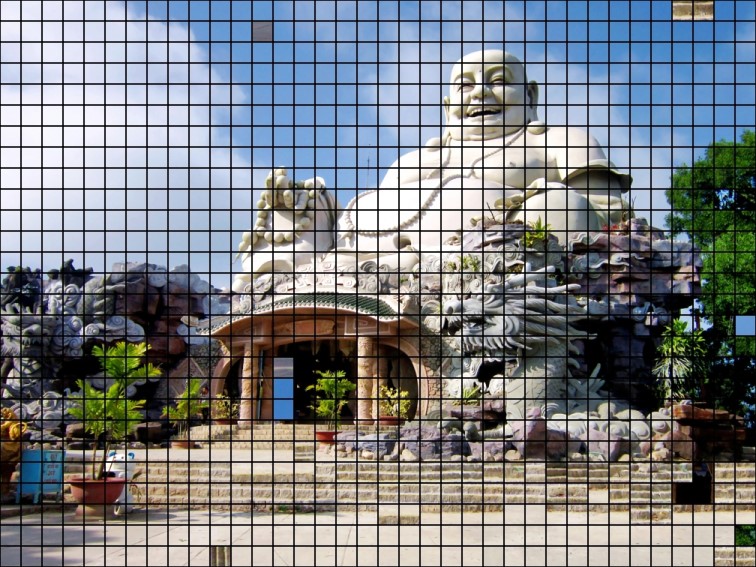}
        \\ \footnotesize{TEN 18.7\%} & \footnotesize{TEN-L 32.4\%} & \footnotesize{DNN-Based E2E 46.4\%} & \footnotesize{Edge2Vec 87.8\%} \\

        \includegraphics[width=0.2111\linewidth]{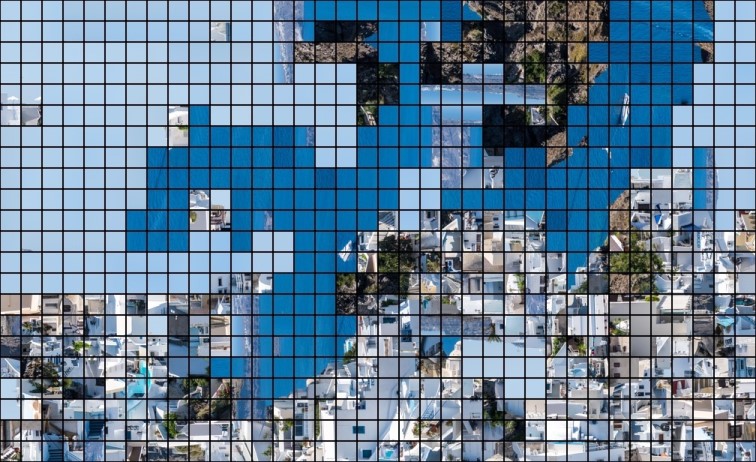} &
        \includegraphics[width=0.2111\linewidth]{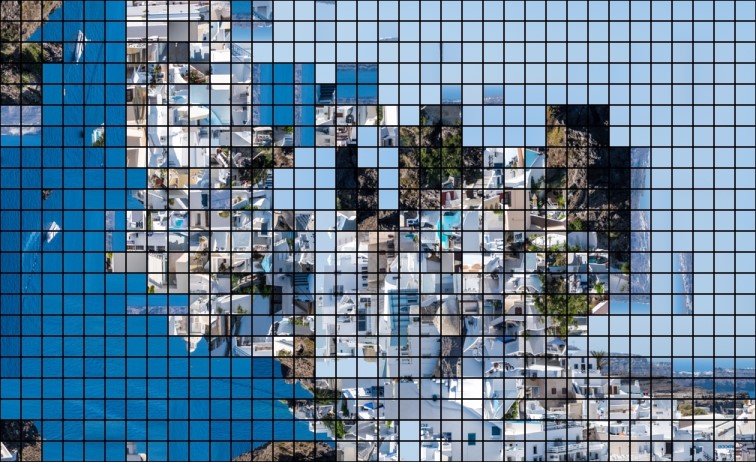} &
        \includegraphics[width=0.2111\linewidth]{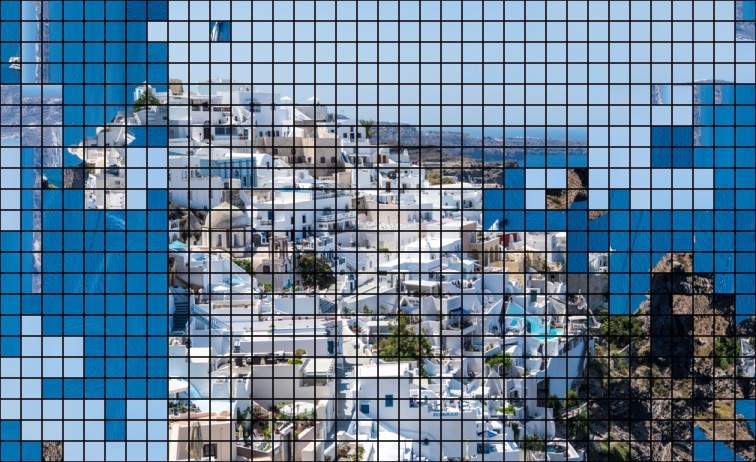} &
        \includegraphics[width=0.2111\linewidth]{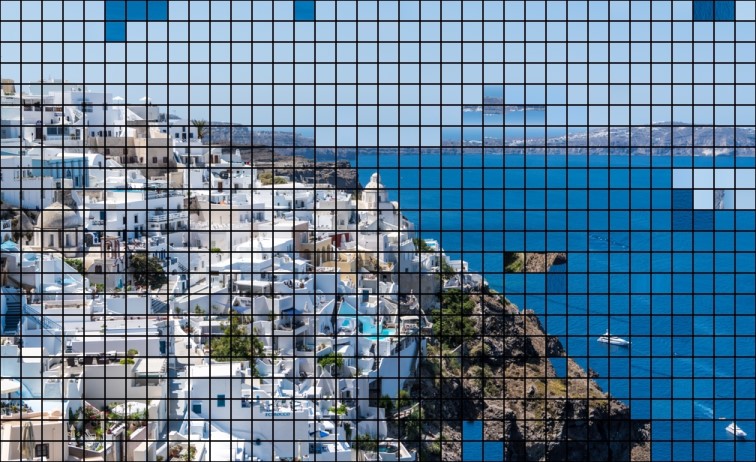}
        \\ \footnotesize{TEN 8.2\%} & \footnotesize{TEN-L 13.1\%} & \footnotesize{DNN-Based E2E 52.5\%} & \footnotesize{Edge2Vec 73.3\%} \\
    \end{tabular}
\caption{Illustration of reconstruction samples due to Gallagher's solver for Type-2 variant; Edge2Vec exhibits superior performance to previous DNN-based embedding and E2E methods.}
\label{fig:reconstruction_comparison}
\end{figure}

\clearpage

\bibliographystyle{unsrt}
\bibliography{egbib}

\begin{thebibliography}{10}

\bibitem{conf/cvpr/ChoAF10}
T.~S. Cho, S.~Avidan, and W.~T. Freeman.
\newblock A probabilistic image jigsaw puzzle solver.
\newblock In {\em Proceedings of the IEEE Conference on Computer Vision and
  Pattern Recognition}, pages 183--190, 2010.

\bibitem{conf/cvpr/PomeranzSB11}
D.~Pomeranz, M.~Shemesh, and O.~Ben-Shahar.
\newblock A fully automated greedy square jigsaw puzzle solver.
\newblock In {\em Proceedings of the IEEE Conference on Computer Vision and
  Pattern Recognition}, pages 9--16, 2011.

\bibitem{gallagher2012jigsaw}
A.~C. Gallagher.
\newblock Jigsaw puzzles with pieces of unknown orientation.
\newblock In {\em Proceedings of the IEEE Conference on Computer Vision and
  Pattern Recognition}, pages 382--389, 2012.

\bibitem{paikin2015solving}
G.~Paikin and A.~Tal.
\newblock Solving multiple square jigsaw puzzles with missing pieces.
\newblock In {\em Proceedings of the IEEE Conference on Computer Vision and
  Pattern Recognition}, pages 4832--4839, 2015.

\bibitem{paixao2018deep}
T.~M. Paixao, R.~F. Berriel, M.~C. Boeres, C.~Badue, A.~F. De~Souza, and
  T.~Oliveira-Santos.
\newblock A deep learning-based compatibility score for reconstruction of
  strip-shredded text documents.
\newblock In {\em Proceedings of the IEEE 31st SIBGRAPI Conference on Graphics,
  Patterns and Images}, pages 87--94, 2018.

\bibitem{SqueezeNet}
N.~I. Forrest, H.~Song, W.M. Matthew, A.~Khalid, J.D. William, and K.~Kurt.
\newblock Squeezenet: Alex{N}et-level accuracy with 50x fewer parameters and
  $<$0.5{MB} model size.
\newblock {\em arXiv:1602.07360}, 2015.

\bibitem{mobilenetv2_2018_cvpr}
M.~Sandler, A.~Howard, M.~Zhu, A.~Zhmoginov, and L.~Chen.
\newblock Mobile{N}et{V}2: Inverted residuals and linear bottlenecks.
\newblock In {\em Proceedings of the {IEEE} Conference on Computer Vision and
  Pattern Recognition}, pages 4510--4520, 2018.

\bibitem{rika2019gecco}
D.~Rika, D.~Sholomon, O.~E. David, and N.~S. Netanyahu.
\newblock A novel hybrid scheme using genetic algorithms and deep learning for
  the reconstruction of {P}ortuguese tile panels.
\newblock In {\em Proceedings of the {ACM} Conference on Genetic and
  Evolutionary Computation}, pages 1319--1327, 2019.

\bibitem{bridger2020cvpr}
D.~Bridger, D.~Danon, and A.~Tal.
\newblock Solving jigsaw puzzles with eroded boundaries.
\newblock In {\em Proceedings of the {IEEE} Conference on Computer Vision and
  Pattern Recognition}, pages 3526--3535, 2020.

\bibitem{ten2022}
D.~Rika, D.~Sholomon, O.~E. David, and N.~S. Netanyahu.
\newblock {TEN}: {T}win {E}mbedding {N}etworks for the {J}igsaw {P}uzzle
  {P}roblem with {E}roded {B}oundaries.
\newblock {\em https://doi.org/10.48550/arxiv.2203.06488}, 2022.

\bibitem{nair2010rectified}
V.~Nair and G.E. Hinton.
\newblock Rectified linear units improve restricted boltzmann machines.
\newblock In {\em Proceedings of the 27th International Conference on Machine
  Learning}, pages 807--814, 2010.

\bibitem{Agustsson_2017_CVPR_Workshops}
E.~Agustsson and R.~Timofte.
\newblock {NTIRE} 2017 challenge on single image super-resolution: Dataset and
  study.
\newblock In {\em Proceedings of the IEEE Conference on Computer Vision and
  Pattern Recognition Workshops}, July 2017.

\bibitem{pirm2018cvpr}
Y.~Blau, R.~Mechrez, R.~Timofte, T.~Michaeli, and L.~Zelnik{-}Manor.
\newblock 2018 {PIRM} challenge on perceptual image super-resolution.
\newblock {\em https://doi.org/10.48550/arxiv.1809.07517}, 2018.

\bibitem{kingma2017adam}
D.~P. Kingma and J.~Ba.
\newblock Adam: A method for stochastic optimization.
\newblock {\em arXiv:1412.6980}, 2017.

\end{thebibliography}

\end{document}